\documentclass[moor]{informs2}

\long\def\hide#1{}

\usepackage{subfigure}
\usepackage{comment}
\usepackage{algorithm}
\usepackage[noend]{algpseudocode}
\usepackage{enumerate}
\usepackage{type1cm}
\usepackage[colorlinks=true,breaklinks=true,bookmarks=true,urlcolor=blue,citecolor=blue,linkcolor=blue,bookmarksopen=false,draft=false]{hyperref}
\usepackage[usenames,dvipsnames,svgnames,table]{xcolor}

\algrenewcommand\alglinenumber[1]{\scriptsize #1:}
\newcommand{\commentsymbol}{//}
\algnewcommand{\LineComment}[1]{\State \commentsymbol #1}

\usepackage{natbib}
 \NatBibNumeric
 \def\bibfont{\small}%
 \def\bibsep{\smallskipamount}%
 \def\bibhang{24pt}%
 \bibpunct[, ]{[}{]}{,}{n}{}{,}%

\newboolean{showcomments}
\setboolean{showcomments}{true}
\newcommand{\raga}[1]{  \ifthenelse{\boolean{showcomments}}
{ \textcolor{Red}{(Raga says:  #1)}} {}  }
\newcommand{\arpita}[1]{\ifthenelse{\boolean{showcomments}}
{ \textcolor{Blue}{(Arpita says: #1)} } {} }

\makeatletter
\newcommand{\subalign}[1]{%
  \vcenter{%
    \Let@ \restore@math@cr \default@tag
    \baselineskip\fontdimen10 \scriptfont\tw@
    \advance\baselineskip\fontdimen12 \scriptfont\tw@
    \lineskip\thr@@\fontdimen8 \scriptfont\thr@@
    \lineskiplimit\lineskip
    \ialign{\hfil$\m@th\scriptstyle##$&$\m@th\scriptstyle{}##$\crcr
      #1\crcr
    }%
  }
}
\makeatother

\def\EMAIL#1{\href{mailto:#1}{#1}}

\TheoremsNumberedThrough
\ECRepeatTheorems

\EquationsNumberedThrough

\MANUSCRIPTNO{XXX}

\begin{document}

\RUNAUTHOR{Gopalakrishnan et al.}

\RUNTITLE{Demand Prediction and Placement Optimization for EV Charging Stations}

\TITLE{Demand Prediction and Placement Optimization for Electric Vehicle Charging Stations}

\ARTICLEAUTHORS{
\AUTHOR{Ragavendran Gopalakrishnan}
\AFF{Xerox Research Centre India, Bangalore, Karnataka 560103,\\ \EMAIL{Ragavendran.Gopalakrishnan@xerox.com}}
\AUTHOR{Arpita Biswas}
\AFF{Xerox Research Centre India, Bangalore, Karnataka 560103,\\ \EMAIL{Arpita.Biswas@xerox.com}}
\AUTHOR{Alefiya Lightwala}
\AFF{Xerox Research Centre India, Bangalore, Karnataka 560103,\\ \EMAIL{Alefiya.Lightwala@xerox.com}}
\AUTHOR{Skanda Vasudevan}
\AFF{Xerox Research Centre India, Bangalore, Karnataka 560103,\\ \EMAIL{Skanda.Vasudevan@xerox.com}}
\AUTHOR{Abhishek Tripathi}
\AFF{Xerox Research Centre India, Bangalore, Karnataka 560103,\\ \EMAIL{Abishek.Tripathi3@xerox.com}}
\AUTHOR{Partha Dutta}
\AFF{Xerox Research Centre India, Bangalore, Karnataka 560103,\\ \EMAIL{Partha.Dutta@xerox.com}}
}

\ABSTRACT{
Due to the environmental impact of fossil fuels and high variability in their prices, there is rising interest in adopting electric vehicles (EVs) by both individuals and governments. Despite the advances in vehicle efficiency and battery capacity, a key hurdle is the inherent interdependence between EV adoption and charging station deployment--EV adoption (and hence, charging demand) increases with the availability of charging stations (operated by service providers) and vice versa. Thus, effective placement of charging stations plays a key role in EV adoption. In the placement problem, given a set of candidate sites, an optimal subset needs to be selected with respect to the concerns of both (a)~the charging station service provider, such as the demand at the candidate sites and the budget for deployment, and (b)~the EV user, such as charging station reachability and short waiting times at the station. This work addresses these concerns, making the following three novel contributions: (i)~a supervised multi-view learning framework using Canonical Correlation Analysis (CCA) for demand prediction at candidate sites, using multiple datasets such as points of interest information, traffic density, and the historical usage at existing charging stations; (ii)~a ``mixed-packing-and-covering'' optimization framework that models competing concerns of the service provider and EV users, which is also extended to optimize government grant allocation to multiple service providers; (iii)~an iterative heuristic to solve these problems by alternately invoking knapsack and set cover algorithms. The performance of the demand prediction model and the placement optimization heuristic are evaluated using real world data. In most cases, and especially when budget is scarce, our heuristic achieves an improvement of 10-20\% over a naive heuristic, both in finding feasible solutions and maximizing demand.
}


\SUBJECTCLASS{Primary: Facilities/equipment planning: location; secondary: Programming: integer}

\maketitle

\section{Introduction.}

\begin{figure}[t]
    \centering
    \includegraphics[width=0.33\textwidth]{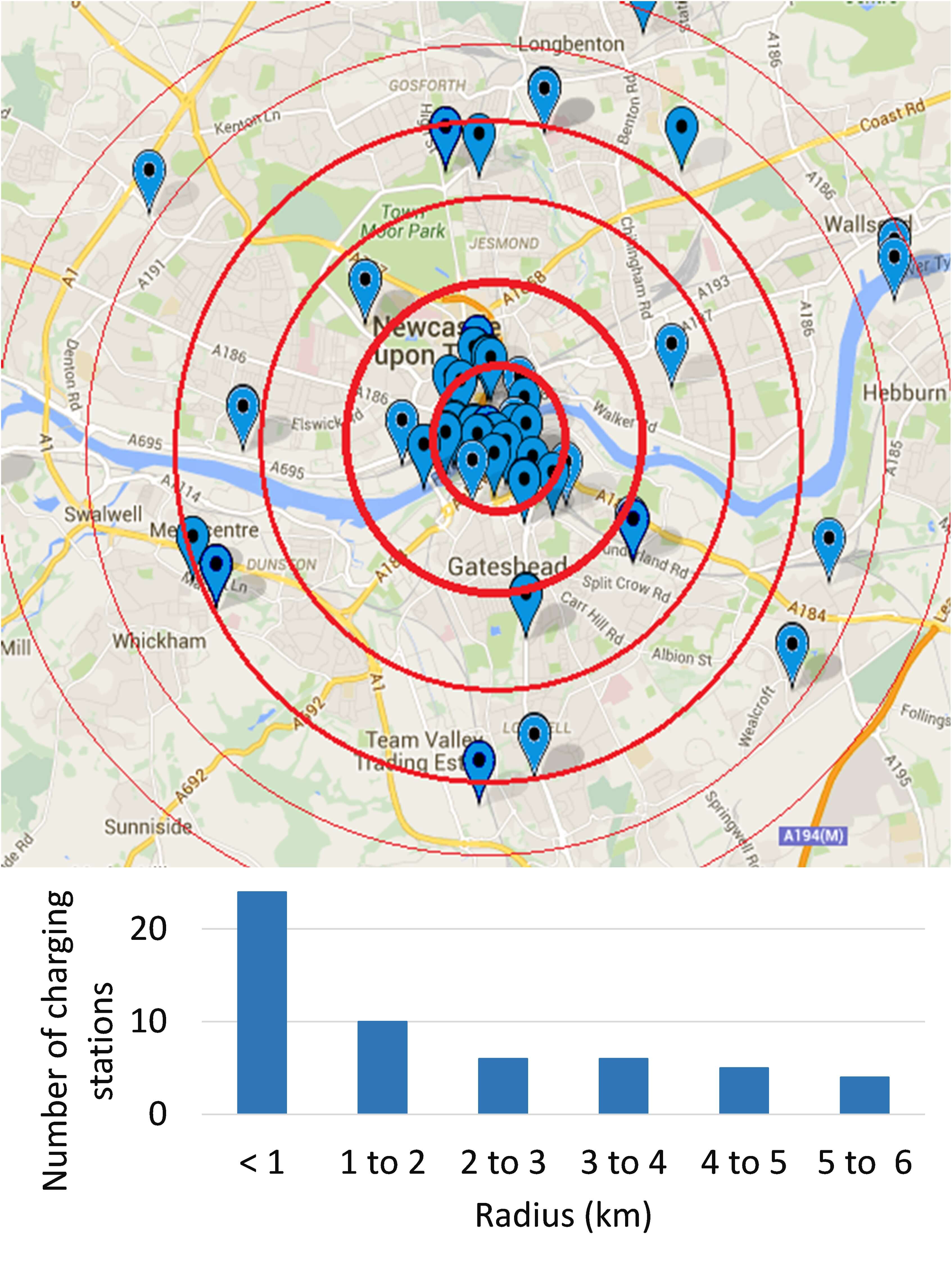}
    \caption{Charging stations in North East England.\label{fig:neuk-map}}
\end{figure}

The environmental impact of fossil fuels and the high variability in their prices have led to rising adoption of Electric Vehicles (EVs), which is supported by ambitious government policies for promoting EVs~\cite{EVUS,EVEU}. Despite the technological advances in vehicle efficiency and battery capacity, a key hurdle in EV adoption is that, barring a few pockets of densely populated areas, the distribution of EV charging stations is sparse in most regions.\footnote{Recent government mandates~\cite{EVEU-Coverage,EVUS-DRIVE} focus specifically on the expansion of refueling infrastructure to cover entire geographical regions.} Because of this, EV owners and potential buyers frequently worry about whether the vehicle will have sufficient charge to travel to their trip destinations or an intermediate charging station. On the other hand, given the high cost of building a charging station and currently low\footnote{Nevertheless, this number is rapidly growing, e.g., in the U. S., EV market share has risen~\cite{EVUS-Growth} from 0.14\% in 2011 to 0.72\% in 2014.} number of EVs, charging station operators would only want to place stations where there is sufficient demand for charging. This results in a situation where the charging station density is concentrated near the city centre and rapidly decrease as we get farther, e.g., Figure~\ref{fig:neuk-map} illustrates this for North East England.

This work is an attempt to break this deadlock by improving the distribution of new charging stations so as to mitigate the range anxiety of EV users. To be effective, such a solution must simultaneously address the concerns of both EV users and charging station operators. Charging station operators are concerned about (a)~\textit{demand}: charging stations must be distributed to serve the maximum total demand for charging, and (b)~\textit{budget}: there is a limited budget available for setting up charging stations. EV users are concerned about (c)~\textit{reachability}: there must be a charging station that is reachable within a short driving distance from most locations, and (d)~\textit{waiting time}: the waiting time to begin charging at charging stations should not be prohibitively large, given that the charging event itself is time-consuming.\footnote{There may be several other concerns for charging station operators and EV users, including alternate interpretations of \textit{reachability}, e.g.,~\cite{Funke15}, but in this work, we restrict our attention to modeling just these four.}

This paper models the above placement problem within a framework that is inspired from the facility location literature \cite{Drezner95}. Given a set of candidate sites for charging stations, e.g., parking lots in a city, the objective is to select an optimal subset of these locations that maximizes the total demand satisfied, subject to the total cost not exceeding the available budget (``packing'' constraint), and given a set of locations of interest, there is at least one charging station within a specified driving range of every location in this set (``covering'' constraint). This ``mixed-packing-and-covering'' formulation (Section~\ref{ssec:prelims-placement-optimization}) covers concerns (a)-(c). In order to take care of (d), a queueing model is used to translate the constraint on waiting time and the charging demand at a location to a constraint on the minimum number of charging slots needed at a charging station at that location, which is then factored into the setup costs in a pre-processing step.

Before tackling the optimization problem itself, it is necessary to predict the demand for charging at candidate sites. Although demand prediction has been extensively studied for planning transportation infrastructure, the EV charging station demand prediction problem is difficult because of sparse deployment of such stations, and lack of sufficient and reliable historic data. Thus, in addition to the limited historical demand data for existing charging stations in other locations, two other types of features of the location which may have a strong impact on the usage are considered. First, nearby Points of Interest (PoI), e.g., shopping malls, institutions, restaurants, hospitals etc., indicate the frequency of visits that could be made by travelers to the given location, and hence, are useful for estimating the demand. Second, traffic density data at nearby road junctions are also useful as features in demand prediction. Using these features, in Section~\ref{sec:demand-prediction} proposes a supervised multi-view learning framework using Canonical Correlation Analysis (CCA) for EV charging station demand prediction.

Once the demands are estimated using the demand prediction algorithm, the next step is to solve the mixed packing and covering problem. We identify a family of heuristics in Section~\ref{mixedpackandcover} that seeks to iteratively find the optimal allocation of the available budget between satisfying the packing and the covering constraints, by alternatingly invoking algorithms that solve reduced knapsack and set-cover subproblems. For example, choosing the well known greedy algorithms for the knapsack and set-cover problems~\cite{Vazirani01} yields one instance in this family.\footnote{Since the primary mixed packing and covering problem, and the associated knapsack and set-cover problems are NP-Hard, it is unlikely that any polynomial-time solution, including this instance, would be optimal.} In Section~\ref{ssec:expt-optimal-placement}, we present results from an experimental evaluation of this instance for the EV charging station placement problem using charging data from the UK. In most cases, and especially when budget is scarce, our heuristic achieves an improvement of 10-20\% over a naive heuristic, both in terms of finding feasible solutions and maximizing demand.

The modeling and heuristic frameworks above can be easily extended to the following alternate scenarios: (a)~incremental placement of charging stations when the budget is progressively released over a period of time, and (b)~when there are multiple charging station operators, each with their respective budgets, and a government agency (subject to its own budget) seeks to optimally allocate grants to incentivize these providers to set up charging stations at selected locations (e.g., where the demand is low).

\subsection{Related work.}

With the recent increase in the adoption of EVs, the charging station placement problem has received significant attention. One of the prerequisites of an effective charging station placement is the availability of estimated charging demand at candidate sites. In this context, prior literature has studied the impact of external information on charging station demand. For example, parking demand is combined with facility location problem in~\cite{Chen13}, and the effect of demographic features is studied using regression models in~\cite{Wagner14}. However, these approaches lack a principled multi-view learning framework to integrate heterogeneous data sets to predict EV charging demand. In the machine learning literature, CCA-based models have been used for multi-variate regression~\cite{Piyush09,Sun11,Klami13}; in this paper, we propose a novel variant of CCA-based multivariate regression.

A useful charging station placement solution needs to take into account the charging demand, budget, coverage, and waiting time at the stations. Existing work has not considered all of these factors simultaneously. For example, placement of charging stations based on predicted demand has been studied in~\cite{Dai13,Klabjan12,Wagner14,Chen13,Lam13,chung2015multi}, but they do not simultaneously consider constraints on budget, coverage, and waiting times. Coverage requirements in charging station placement has been modeled as path cover in~\cite{Funke15}, and reachability cover in~\cite{Funke13} over the city road network, but they do not consider the demand satisfied by the deployment and the waiting time faced by the users.

The optimization problem considered in this work contains both packing and covering constraints, with integer decision variables. Mixed packing and covering problems have been studied for fractional decision variables~\cite{Young01,Azar13}, as well as for integer decision variables~\cite{Chakaravarthy13,Mukherjee15}. However, this work introduces an optimization problem with a knapsack packing constraint and a set covering constraint, which has not been studied earlier.

\section{Demand prediction at candidate sites.}\label{sec:demand-prediction}

Understanding the EV charging demand is key to optimal placement of charging stations. Since the distribution of charging stations is typically sparse, there is insufficient data to efficiently model the demand. Thus, it becomes important to model the relationship of EV charging demand with auxiliary data such as Point of Interest (PoI) and traffic density. Such relationships help in better understanding not only the charging demand, but also the factors behind it. Additionally, the expected demand at a candidate site where no historical demand is known, can be predicted based on external factors. A multi-view learning framework based on Canonical Correlation Analysis (CCA) is used to jointly model the charging demand data and other external data sources, and to predict the demand at new candidate sites given the external factors. Next, we briefly describe CCA and then present the CCA-based regression framework.

\subsubsection{Canonical correlation analysis.}

Canonical correlation analysis is a classical method to find a linear relationship between two sets of random variables. Let $X$ and $Y$ be data matrices of size $n \times d_x$ and $n \times d_y$ respectively, where each row corresponds to a realization of the random variables. CCA finds linear projections $u_x \in \mathbb{R}^{d_x}$ and $u_y \in \mathbb{R}^{d_y}$ such that their correlation, $corr(Xu_x,Yu_y)$, is maximized. It can be shown that maximizing correlation is equivalent to minimizing $\| Xu_x - Yu_y\|^2$ subject to $\| Xu_x\|^2 =1$ and $\| Yu_y\|^2 = 1$. A set of $k \leq min(rank(X),rank(Y))$ linear projections can be computed by solving a generalized eigenvalue problem;~\cite{Hardoon04} provides a detailed review of classical CCA.

\subsection{Multi-view regression using CCA.}

In this section, a multi-view learning based regression framework for the charging demand prediction at candidate sites is presented. The regression framework uses CCA in a supervised setting as a multivariate regression tool. CCA is, in fact, closely related to multivariate linear regression;~\cite{Borga01,Sun11} present least-square formulations. CCA-based approaches model the statistical dependence between two or more data sets and assume a single latent factor to describe information shared between all datasets. In a supervised setting such as multivariate regression task, the output variable is considered as one of the random variables and CCA models the statistical dependence between the output variable (dependent) and a covariate (independent variable). However, when there are more than two datasets (output variable and multiple covariates), modeling shared information may not always be a good choice, since it would discard dataset-specific information which may be crucial to predicting the output variable.

To address this deficiency, a multi-view learning regression framework is proposed when there are $n>2$ covariates, all of which can be used to predict the output variable. The idea is to model the statistical dependence of the output variable with each of the covariates separately and then learn a weighted combination for each such model that maximizes the prediction performance. We call this model Multiple Dependent Regression (MDR), which can be formulated as the ensemble regression function $\mathbf{f} = \sum_i w_i \mathbf{f_i}$, where $\sum_i w_i=1$. Here, $\mathbf{f_i}$ is a regression function learned from covariate $X_i$ to the output variable $Y$, and $w_i$ are the weights denoting the importance of covariate $X_i$ in the regression task. The regression function $\mathbf{f_i}$ for covariate $X_i$ can be learnt according to a CCA-based regression, given by $\argmin_{U_x^i,U_y^i} \|X_iU_x^i - YU_y^i\|^2$, where $U_x^i$ and $U_y^i$ are $k$-dimensional linear projections for each (covariate, output variable) pair $(X_i,Y)$. Hence, MDR is a two step framework of multiple regression:
\begin{enumerate}[(1)]
\item For each covariate $X_i$, apply CCA for $X_i$ and $Y$, and compute the prediction error $e_i$.
\item For each $i$, compute $w_i = \frac{1}{n-1}(1 - \frac{e_i}{\sum_j e_j})$.
\end{enumerate}
The values of parameters $U_x, U_y, w_i$ are learnt from the training data as described in the steps above. The model thus trained can be used to predict the output variable given the input covariates. In our case, the task is to predict the demand of EV charging at a new candidate location given the covariates as explained earlier.

MDR is analogous to ensemble learning approaches where multiple learning models are trained to solve a given problem. In our case, we train multiple CCA-based regression models in the first step and learn a weighted combination of models based on training error in the second step. The weights $w_i$ explain the importance of each covariate in the prediction task. This procedure is flexible--any CCA-based solution can be used in Step~(1). In this work, we use a Bayesian solution with group-wise sparsity proposed in~\cite{Klami13}.

By learning a separate CCA model for each covariate, MDR ensures that dataset-specific information of any covariate (with respect to other covariates) is not ignored if it is shared with the output variable. MDR provides an intuitive, flexible, and yet, a simple approach to multi-view regression. It can be easily extended to non-linear projections using non-linear solutions to CCA.

\subsection{Experimental results for demand prediction.}\label{ssec:expt-demand-prediction}

MDR is evaluated on EV charging data obtained from $252$ public charging points in North East England through UK's Plugged-In-Places program~\cite{PiP}. The location of charging points was obtained from UK's National Charge Point Registry data~\cite{NCPR}. PoI information is extracted from OpenStreetMap~\cite{Haklay08} API for $11$ categories: sustenance, education, transportation, financial, healthcare, entertainment, sports, gardens, places of worship, shops and public buildings. Finally, we use traffic data are for each junction-to-junction link on major road networks, provided by \cite{TrafficData}.

The following data matrices are created: (i)~$Y \in \mathbb{R}^{252 \times 24}$ represents hour-wise charging demand for each of $252$ charging points, where $Y[i,j]$ is the average energy consumed at charging point $i$ in hour $j$; (ii)~$X_1 \in \mathbb{R}^{252 \times 11}$ represents the PoIs, where $X_1[i,j]$ is the frequency of PoI category $j$ within a radius of $500$ meters around charging point $i$; (iii)~$X_2 \in \mathbb{R}^{252 \times 5}$ represents the traffic densities at the $5$ nearest traffic junctions to each charging point; and (iv)~$X_3 \in \mathbb{R}^{252 \times 5}$ represents the charging demands at the $5$ nearest charging points to each charging point. Since charging data is sparse during early mornings and late nights, only data from 07:00 hrs. till 23:00 hrs. is used.

\begin{figure}[ht]
    \centering
    \includegraphics[width=0.5\textwidth]{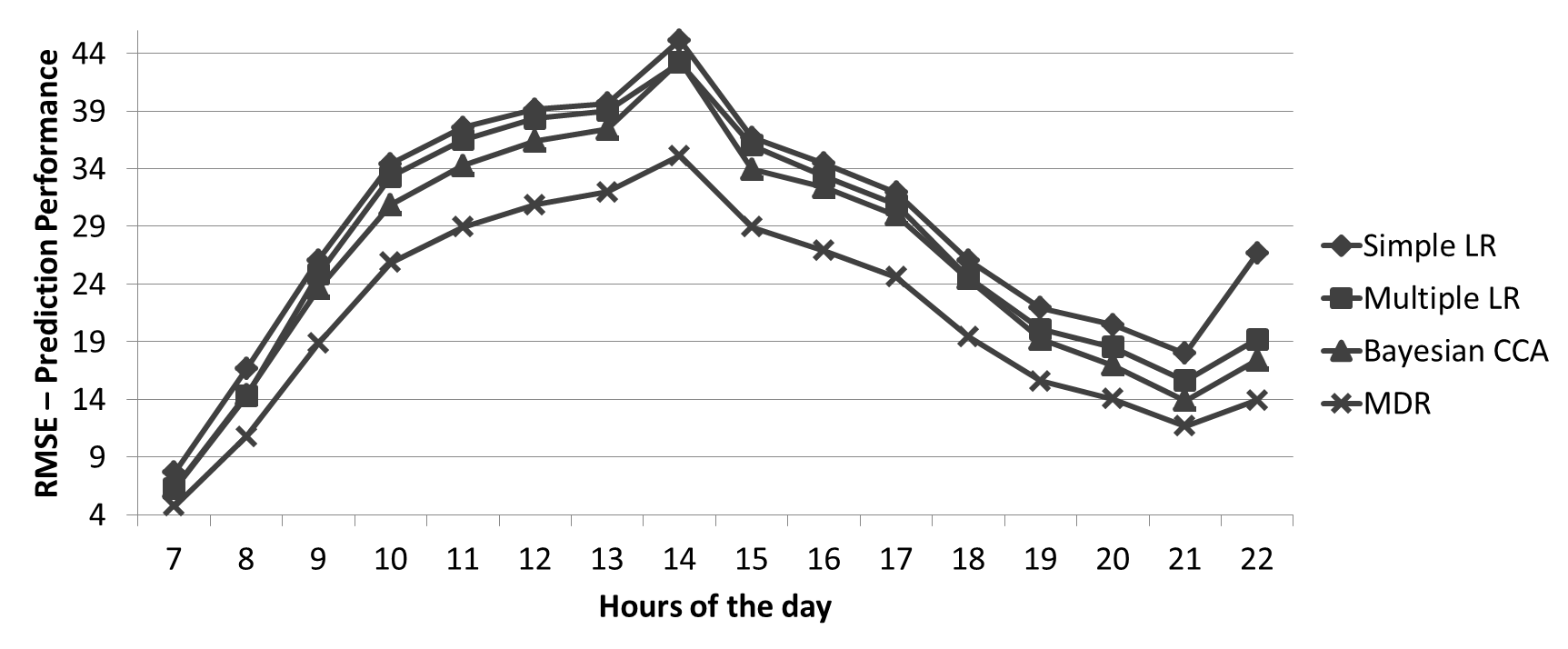}
    \caption{Comparison of MDR against baselines.\label{fig:MDR}}
\end{figure}

MDR is evaluated in a leave-one-out cross-validation manner by training on all but one instance (charging point) and testing on the left-out instance. Prediction performance is measured as the average Root Mean Square Error (RMSE) over all test instances.  MDR is compared against (i)~Linear Regression(LR) - all covariates are concatenated feature-wise similar to~\cite{Wagner14}, (ii)~Multiple Linear Regression (MLR) - weighted sum of multiple LR analogous to MDR, and (iii)~Bayesian CCA (bCCA) - group factor analysis~\cite{Virtanen12}. Figure~\ref{fig:MDR} shows the average error for each hour of the day over all charging stations. It can be observed that MDR clearly outperforms the other three baselines giving an overall reduction in error by 27\% compared to simple LR, 21\% compared to multiple LR and 18\% compared to bCCA.

\section{Optimal selection from candidate sites for placing charging stations.}\label{sec:placement-optimization}

In this section, we explain the mathematical model for the charging station placement problem, and detail the methods to solve the resulting mixed packing and covering optimization problem.

\subsection{Preliminaries.}\label{ssec:prelims-placement-optimization}
Let $\mathcal{L}=\{1,\ldots,|\mathcal{L}|\}$ denote the set of all candidate sites for placing a charging station. Let $r\in\mathbb{R}_+$ denote the desired ``reachability radius,'' that is, the maximum distance to be travelled in order to reach a charging station. Let $\mathcal{I}$ denote the set of all locations of interest that are desired to be ``covered,'' that is, lie within the reachability radius from at least one charging station. Let $B$ denote the total budget available.

For each candidate site $i\in\mathcal{L}$: $x_i\in\{0,1\}$, $d_i\in\mathbb{R}_+$, and $c_i\in\mathbb{R}_+$ denote, respectively, whether a charging station is placed, the demand for service, and the cost of setting up a charging station at $i$. $S_i^r\subseteq\mathcal{I}$ is the \textit{cover set} of $i$, that is, the set of locations of interest that would be ``covered'' (within a driving distance of $r$) if a charging station is placed at $i$.

\subsubsection{Optimizing for demand.}
For a given reachability radius $r$, set of candidate sites $\mathcal{L}$, and set of locations of interest $\mathcal{I}$, the cover sets $\left\{S_i^r\right\}_{i\in\mathcal{L}}$ can be precomputed. Given predicted demand and charging duration data, a queueing model~\cite{Kleinrock75} can estimate the number of charging slots necessary at each candidate site to meet a specified constraint on waiting times, which is then used to precompute the costs $\left\{c_i\right\}_{i\in\mathcal{L}}$. (See Appendix \ref{appendix-cost-queueing} for details.) Then, the optimization problem to be solved is the Mixed Pack \& Cover (MPC) problem shown in Figure \ref{fig:eqn-MPC}.\footnote{The total demand satisfied by installing charging stations at a subset of candidate sites may be less than the sum of the predicted demands at those sites due to overlapping reachability regions. We ignore this effect in order to keep the optimization problem simple.}

\subsubsection{Approximation algorithm for a transformed problem.}
\begin{figure*}[!t]
    \centering
    \subfigure[\scriptsize{Mixed Pack \& Cover (MPC)}]{\label{fig:eqn-MPC}\includegraphics[width=0.32\textwidth]{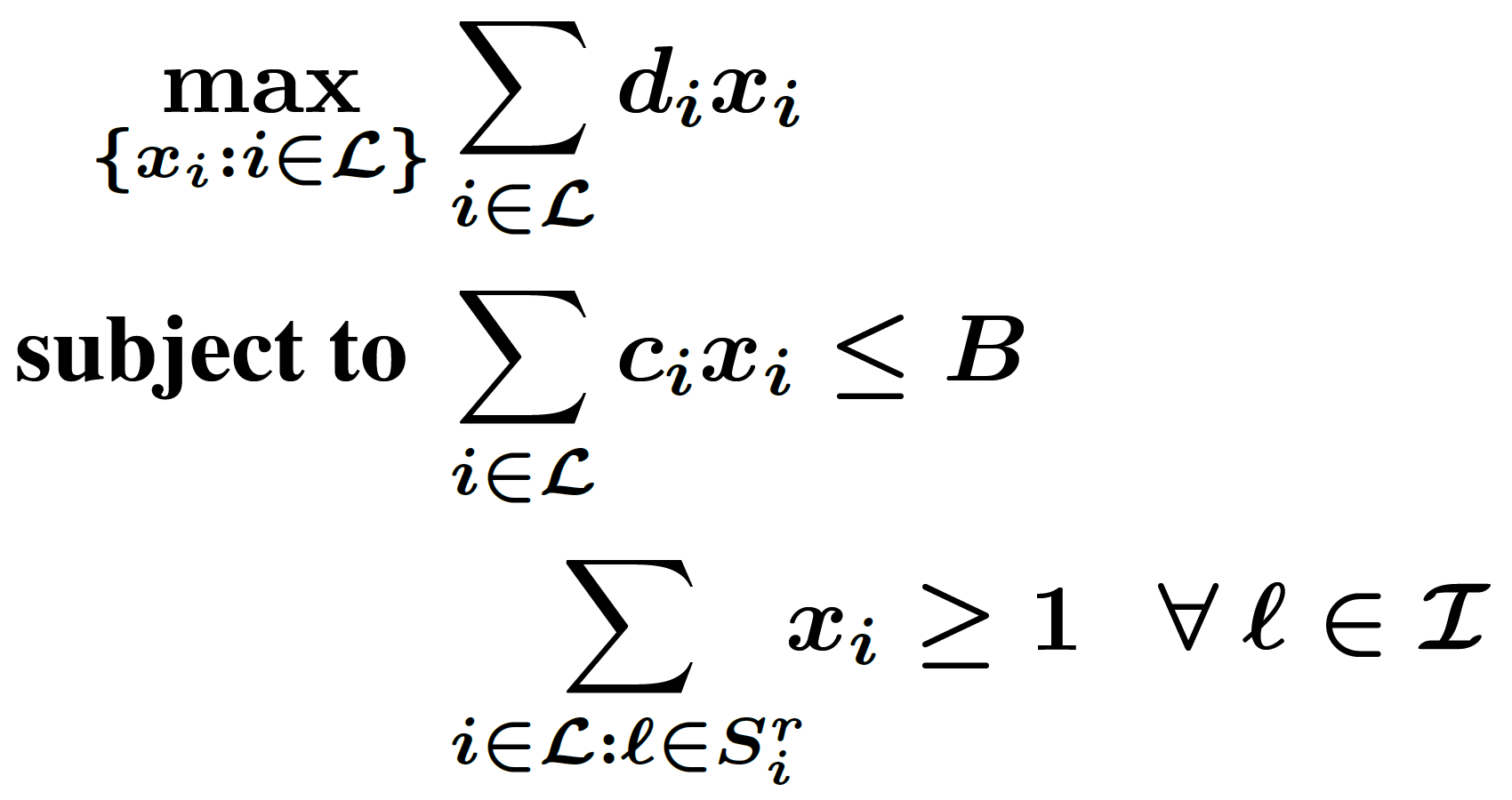}}
    \hfill
    \subfigure[\scriptsize{Knapsack (KP)}]{\label{fig:eqn-KP}\includegraphics[width=0.32\textwidth]{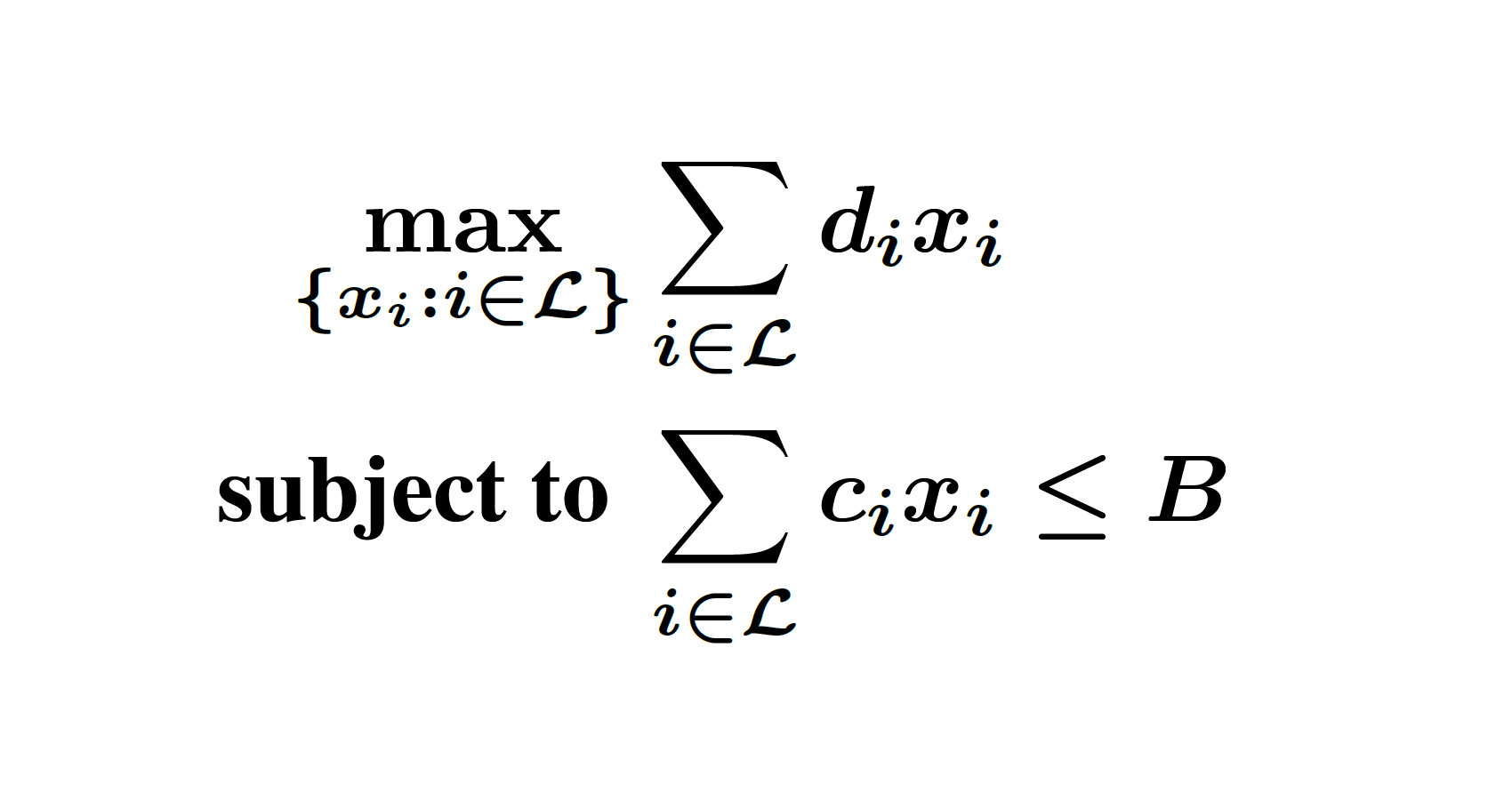}}
    \hfill
    \subfigure[\scriptsize{Set-Cover (SC)}]{\label{fig:eqn-SC}\includegraphics[width=0.32\textwidth]{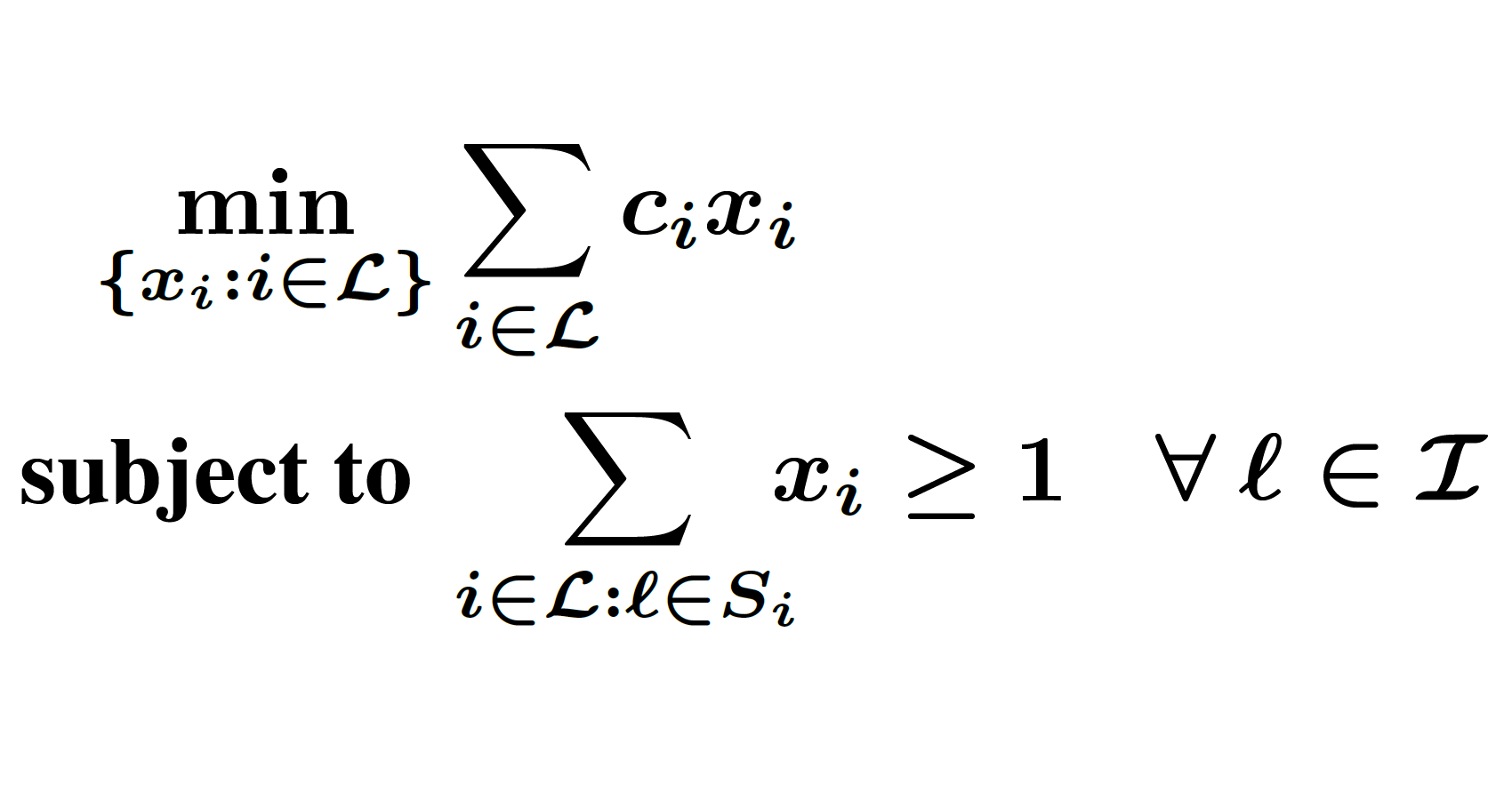}}
    \\
    \subfigure[\scriptsize{Min-Knapsack (MinKP)}]{\label{fig:eqn-MinKP}\includegraphics[width=0.32\textwidth]{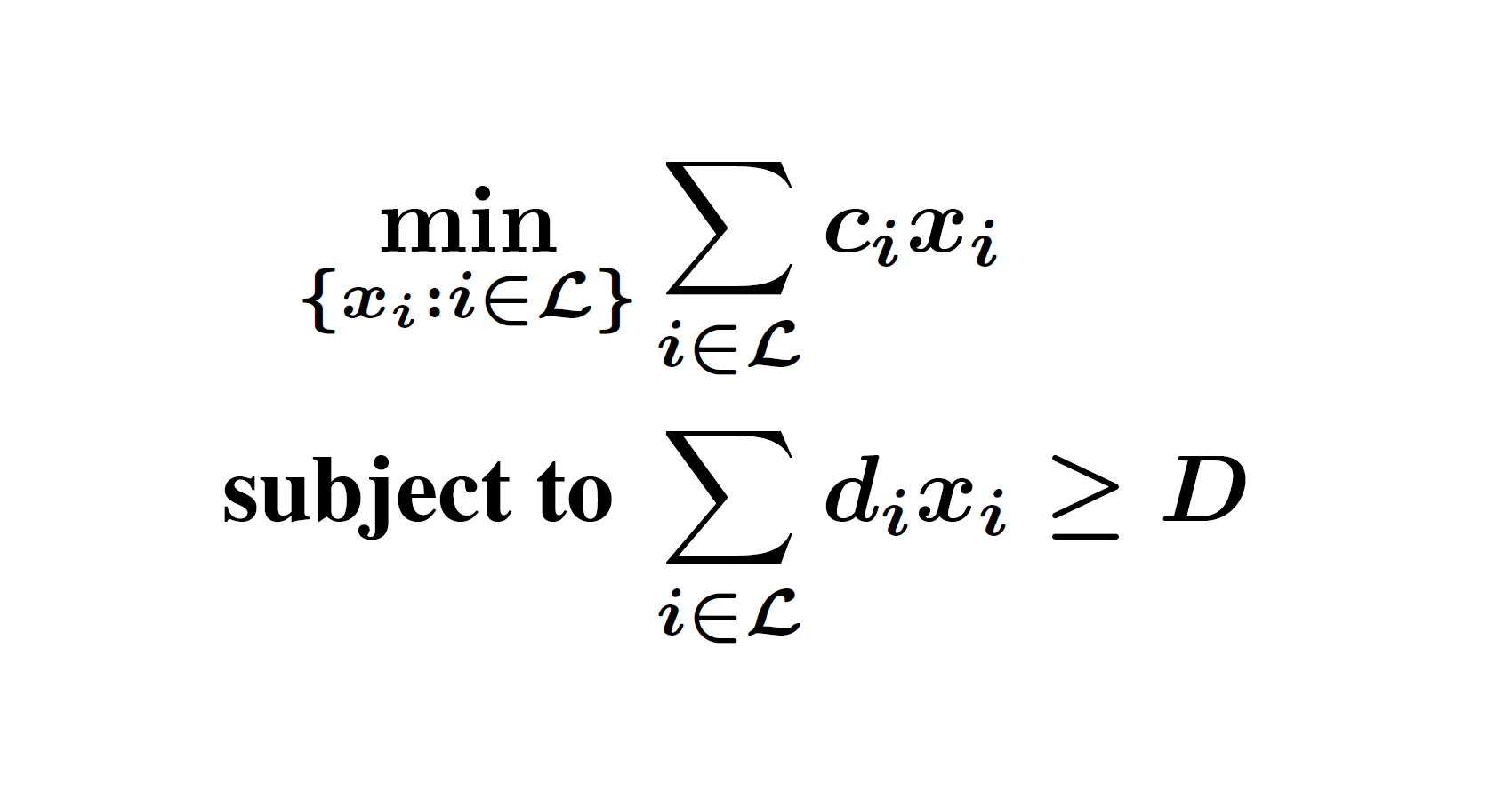}}
    \hspace{0.1in}
    \subfigure[\scriptsize{Demand \& Set-Cover (DSC)}]{\label{fig:eqn-DSC}\includegraphics[width=0.32\textwidth]{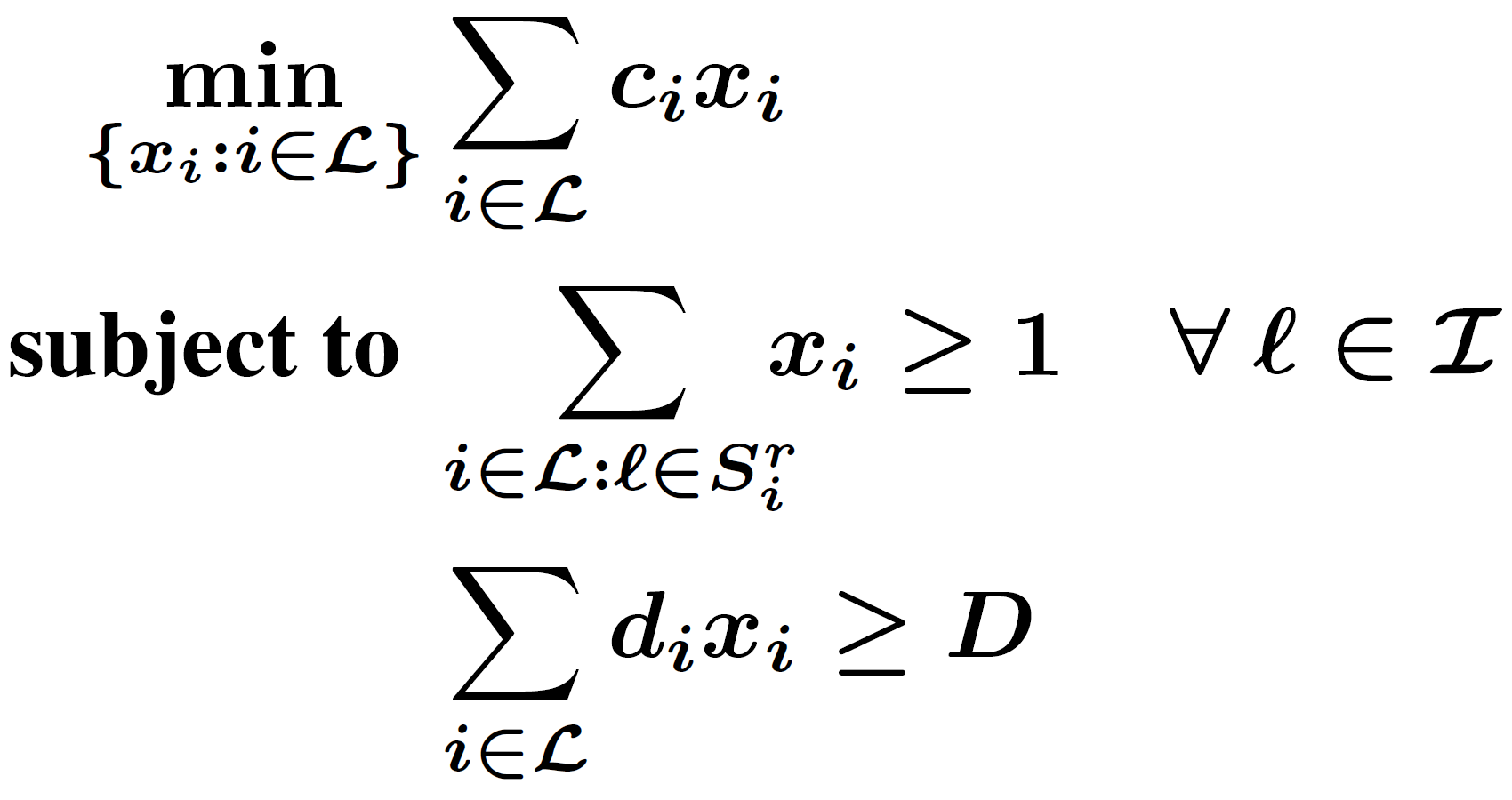}}
    \caption{Important optimization problems.\label{fig:eqn-opt-probs}}
\end{figure*}

MPC (Figure~\ref{fig:eqn-MPC}) is NP-Hard as it contains as a special case, the set-cover problem~\cite{Vazirani01}. For a closely related pure covering problem called Demand \& Set-Cover (DSC, Figure~\ref{fig:eqn-DSC}), an approximation algorithm is described next.

DSC has as subproblems, two well-known NP-Hard problems, Set-Cover (SC)~\cite{Vazirani01} and Minimization Knapsack (MinKP)~\cite{Csirik91}, which are given in Figures~\ref{fig:eqn-SC} and~\ref{fig:eqn-MinKP}, respectively. Consider, an $f_{sc}$ factor approximation algorithm $A_{sc}$ for set cover and an $f_{kp}$ factor approximation algorithm $A_{kp}$ for minimization knapsack. Let $\mathcal{L}_{sc}$ and $\mathcal{L}_{kp}$ (subsets of $\mathcal{L}$) be the solutions returned by $A_{sc}$ and $A_{kp}$, respectively. Consider algorithm $A_{dsc}$ for DSC that uses $A_{sc}$ and $A_{kp}$ as subroutines, and returns the union of the solutions returned by the two subroutines, that is, $\mathcal{L}_{sc} \cup \mathcal{L}_{kp}$, as a solution for DSC. The following lemma (proved in Appendix \ref{appendix-approx-algo}) establishes that $A_{dsc}$ is an $(f_{sc}+f_{kp})$ factor approximation algorithm for DSC.
\begin{lemma}\label{lemma:appendix-approx-algo}
$A_{dsc}$ returns a feasible solution for DSC whose cost is at most $(f_{sc} + f_{kp})$ times the optimal cost.
\end{lemma}
The heuristic framework for MPC presented next is inspired by $A_{dsc}$ and the observation that, similar to DSC, a solution for MPC can be obtained by combining set cover and knapsack algorithms.

\subsection{Solving the mixed-pack-and-cover problem.}\label{mixedpackandcover}
Our goal is to introduce a general methodology for algorithms to solve MPC (Figure \ref{fig:eqn-MPC}), which is NP-Hard, by breaking it down into the following two subproblems:
\begin{enumerate}
\item \textit{The Knapsack Problem (KP):} This is the packing problem in Figure \ref{fig:eqn-KP}. Let $\mbox{\texttt{KP}}\left(\mathcal{L},\{d_i\}_{i\in\mathcal{L}},\{c_i\}_{i\in\mathcal{L}},B\right)$ denote any algorithm that solves this problem.
\item \textit{The Set-Cover Problem (SC):} This is the covering problem in Figure \ref{fig:eqn-SC}. Let $\mbox{\texttt{SC}}\left(\mathcal{L},\mathcal{I},\{c_i\}_{i\in\mathcal{L}},\{S_i\}_{i\in\mathcal{L}}\right)$ denote any algorithm that solves this problem.
\end{enumerate}
The algorithm \texttt{KP()} (respectively, \texttt{SC()}) need not be optimal, but we require that it be nontrivial in the sense that its solution must be \textit{maximal} (respectively, \textit{minimal}). That is, it should not be possible to add (remove) an item and still satisfy the packing (covering) constraint(s).

\subsubsection{The iterative-pack-and-cover (IPAC) framework.}\label{iter-pack-cover}
The available budget could be used very differently if the problem had been a pure packing problem (where maximizing the demand is the only concern) or a pure covering problem (where satisfying the covering constraints is the only concern). Thus, a good solution to the mixed packing and covering problem should achieve an appropriate balance by dividing the available budget between these two concerns. The core idea behind the Iterative Pack And Cover (IPAC) framework is to iteratively search for such an optimal split.

In each iteration, the budget $B$ can be thought of as comprising of three parts: $B^{p}$, $B^{c}$, and some excess (due to the integrality constraint). Here, $B^{c}$ is the portion of the budget used by a solution to a covering problem, and $B^{p}$ is the portion of the budget used by a solution to a packing problem when constrained by a reduced budget of $B-B^{c}$. A check is performed using \texttt{SC()} to determine whether the remaining budget $B-B^{p}$ is sufficient to satisfy the covering constraints left unsatisfied by the solution to the packing problem. Starting with $B^{c}=0$ (pure packing) and the corresponding solution from \texttt{KP()}, during each iteration, $B^{c}$ is increased until the covering check passes, at which point the solutions of the packing and covering problems obtained in the last iteration are merged.\footnote{As an extension, note that $B^{c}$ need not necessarily increase in each iteration, as long as it can be shown that the procedure will terminate in finite time.} The resulting solution is guaranteed to be a feasible solution to the MPC problem (Figure \ref{fig:eqn-MPC}). Optionally, \texttt{KP()} can be invoked one final time to use up any remaining portion of the budget. In the worst case, the iterations continue until $B^{c}=B$, at which point it becomes a pure covering problem, and if the covering check still fails, then IPAC fails to find a feasible solution to MPC. (But if this happens, it cannot necessarily be concluded that MPC is infeasible, unless \texttt{SC()} is optimal, which is unlikely in polynomial time since the set-cover problem is NP-Hard.)

The above framework is quite generic and encompasses a large class of algorithms. Next, we describe a particular instantiation of the framework which runs in polynomial time. The corresponding pseudocode is detailed in Algorithm \ref{alg:iter-pack-cover}.
\begin{enumerate}
\item Set $B^{c}=0$ and invoke \texttt{KP()} to solve the pure packing problem. Let $B^{p}$ be the portion of the budget used by the chosen items. $B^{free}=B-B^{p}$ is the remainder.
\item Update $B^{c}$ to be the minimum budget required to satisfy the unsatisfied covering constraints, computed by invoking \texttt{SC()} to solve the residual covering problem.
\item Repeat the following two steps until either $B^{c}\leq B^{free}$ or $B^{c} > B$:
    \begin{enumerate}
    \item \textit{Packing:} In response to the reduced budget $B-B^{c}$, remove one or more chosen items. In order to do so, use a ranking method, \texttt{RANK()}, that ranks the currently chosen items according to some measure of their importance. Keep removing items that are least important until the remaining items satisfy the reduced budget. Let $B^{p}$ be the portion of the budget used. $B^{free}=B-B^{p}$ is the new remainder.
    \item \textit{Covering:} Since the removal of some items in the previous step might have resulted in more unsatisfied covering constraints, invoke \texttt{SC()} on the new residual covering problem and update $B^{c}$ accordingly.
    \end{enumerate}
\item If $B^{c} > B$, then a feasible solution cannot be found. Otherwise, $B^{c}\leq B^{free}$, and so, add the items from the solution to the last instance of \texttt{SC()} to the remaining chosen items to obtain a feasible solution.
\item Invoke \texttt{KP()} to fill any unused portion of the budget using unallocated items.
\end{enumerate}

\begin{algorithm}[!ht]
\caption{\texttt{IPAC}\label{alg:iter-pack-cover}}
\begin{algorithmic}[1]
    \scriptsize
    \State \textbf{Input:} $\mathcal{L}$, $\mathcal{I}$, $\{d_i\}_{i\in\mathcal{L}}$, $\{c_i\}_{i\in\mathcal{L}}$, $B$, $\{S_i^r\}_{i\in\mathcal{L}}, \mathtt{KP}(), \mathtt{SC}(), \mathtt{RANK}()$
    \State \textbf{Output:} $\{x_i\}_{i\in\mathcal{L}}$
    \item[]
    \State $\{x_i\}_{i\in\mathcal{L}} \leftarrow \mathtt{KP}(\mathcal{L}, \{d_i\}_{i\in\mathcal{L}}, \{c_i\}_{i\in\mathcal{L}},B)$;
    \State $\mathcal{L}^{p} \leftarrow \{i\in\mathcal{L}\ :\ x_i = 1\}$; $B^{p} \leftarrow \sum_{i\in\mathcal{L}}c_ix_i$;
    \State $\mathcal{I}^{p} \leftarrow \bigcup_{i\in\mathcal{L}^{p}}S_i^r$;
    \LineComment{$\ \mathcal{I}^{p}$ $\leftarrow$ elements of $\mathcal{I}$ covered by $\mathcal{L}^{p}$.}
    \State $\{x^{c}_i\}_{i\in\mathcal{L}\backslash\mathcal{L}^{p}} \leftarrow \mathtt{SC}(\mathcal{L}\backslash\mathcal{L}^{p}$, $\mathcal{I}\backslash\mathcal{I}^{p}$, $\{c_i\}_{i\in\mathcal{L}\backslash\mathcal{L}^{p}}$, $\{S_i^r\}_{i\in\mathcal{L}\backslash\mathcal{L}^{p}})$;
    \State $B^{c} \leftarrow \sum_{i\in\mathcal{L}\backslash\mathcal{L}^{p}}c_ix^{c}_i$; $\ \ B^{free} \leftarrow B - B^{p}$;
    \While{($B^{c} > B^{free}$ and $B^{c} \leq B$)}
      	\LineComment{$\  \rho$ is $\mathcal{L}^{p}$ in increasing order of importance.}
        \State $\rho \leftarrow \mathtt{RANK}(\mathcal{L}^{p}$, $\{d_i\}_{i\in\mathcal{L}^{p}}$, $\{c_i\}_{i\in\mathcal{L}^{p}}$, $\{S_i^r\}_{i\in\mathcal{L}^{p}})$;
        \LineComment{ Remove least important items.}
        \State $j \leftarrow 0$;
        \While{$B^{free} < B^{c}$}
           \State $j \leftarrow j + 1$; $x_{\rho[j]} \leftarrow 0$; $B^{free} \leftarrow B^{free} + c_{\rho[j]}$;
        \EndWhile
        \LineComment{ Recompute quantities.}
        \State $\mathcal{L}^{p} \leftarrow \{i\in\mathcal{L}\ :\ x_i = 1\}$; $\mathcal{I}^{p} \leftarrow \bigcup_{i\in\mathcal{L}^{p}}S_i^r$;
        \State $\{x^{c}_i\}_{i\in\mathcal{L}\backslash\mathcal{L}^{p}} \leftarrow \mathtt{SC}(\mathcal{L}\backslash\mathcal{L}^{p}$, $\mathcal{I}\backslash\mathcal{I}^{p}$, $\{c_i\}_{i\in\mathcal{L}\backslash\mathcal{L}^{p}}$, $\{S_i^r\}_{i\in\mathcal{L}\backslash\mathcal{L}^{p}})$;
        \State $B^{c} \leftarrow \sum_{i\in\mathcal{L}\backslash\mathcal{L}^{p}}c_ix^{c}_i$;
    \EndWhile
    \If{$B^{c} > B$}  EXIT;  \EndIf
    \LineComment{ Add required items to knapsack.}
    \For{$i$ in $\mathcal{L}\backslash\mathcal{L}^{p}$}
         $x_i \leftarrow x_i + x^{c}_i$;
    \EndFor
    \LineComment{ Add more items if possible.}
    \State $\mathcal{L}^{p} \leftarrow \{i\in\mathcal{L}\ :\ x_i = 1\}$; $B^{p} \leftarrow \sum_{i\in\mathcal{L}}c_ix_i$;
    \State $\{x_i\}_{i\in\mathcal{L}\backslash\mathcal{L}^{p}} \leftarrow \mathtt{KP}(\mathcal{L}\backslash\mathcal{L}^{p}$, $\{d_i\}_{i\in\mathcal{L}\backslash\mathcal{L}^{p}}$, $\{c_i\}_{i\in\mathcal{L}\backslash\mathcal{L}^{p}}$, $B-B^{p})$;
\end{algorithmic}
\end{algorithm}

\subsubsection{The \texttt{RANK()} function.}\label{rank}
The effectiveness of IPAC depends on the choice of methods for \texttt{KP()}, \texttt{SC()}, and \texttt{RANK()}. There are several choices in the literature for the first two \cite{Vazirani01}, so we briefly discuss one possible choice for the third. A general observation is that an item $i\in\mathcal{L}$ is more important or desirable if its demand $d_i$ is high, or its cost $c_i$ is low, or if the number of elements it covers $|S_i|$ is high. Based on this, a viable candidate for $\mbox{\texttt{RANK}}\left(\mathcal{L},\{d_i\}_{i\in\mathcal{L}},\{c_i\}_{i\in\mathcal{L}},\{S_i\}_{i\in\mathcal{L}}\right)$ would be a method that ranks items in increasing order according to the value $v_i=\left(\frac{d_i}{\sum_{j\in \mathcal{L}} d_j} + \frac{|S_i|}{|\mathcal{I}|}\right)/c_i$, where $\mathcal{I}=\bigcup_{i\in\mathcal{L}}S_i$ denotes the set of all elements covered by items in $\mathcal{L}$.

\subsubsection{Computing charging station costs.}\label{appendix-cost-queueing}
In this section, we briefly address estimating the setup costs $c_i$, which depend on (i)~the number of charging slots necessary at $i$ to satisfy any desired SLA on waiting times, and (ii)~per-unit setup cost, which can include infrastructure as well as land/licensing costs. Since data on the per-unit costs are generally available, we focus on (i), for which we model candidate charging stations as a multi-server queue.

One possible way to model each candidate charging station at location $i$ in order to estimate its size (and hence cost) is using a multi-server queue that follows an $M$/$M$/$N_i$ discipline, where $N_i\in\mathbb{Z}_+$ is the number of charging slots to be set up. Customers arrive into the queue according to a Poisson process with rate $\lambda_i$ per time unit. The service time for each customer is exponentially distributed with mean $1/\mu_i$ time units. $\lambda_i$ can be derived from the demand $d_i$, whereas $\mu_i$ can be derived from existing data on the average customer service time in nearby existing charging stations. The average time a customer waits before service is then given by $\mathbb{E}[W] = \frac{\mbox{\texttt{ErlC}}\left(N_i,\frac{\lambda_i}{\mu_i}\right)}{N_i\mu_i-\lambda_i}$, where $\mbox{\texttt{ErlC}}(N,\rho) = \left(\frac{\rho^N}{N!}\frac{N}{N-\rho}\right) / \left(\sum_{j=0}^{N-1}\frac{\rho^j}{j!}+\frac{\rho^N}{N!}\frac{N}{N-\rho}\right)$~\cite{Kleinrock75}. Suppose the SLA to be met is given by $\mathbb{E}[W] \leq t_i$. Then, because it is well-known~\cite{Kleinrock75} that $\mathbb{E}[W]$ is a decreasing function of $N_i$, we simply choose the smallest $N_i$ (e.g., using binary search) such that the SLA is satisfied.

\subsubsection{Optimizing for reachability.}\label{ssec:reachability}
In addition to maximizing the demand for a given reachability radius, it may also be desirable to minimize the reachability radius itself. Without loss of generality, we assume that $r\in[R^{\min},R^{\max}]$, where $R^{\min} = \max_{\ell\in\mathcal{I}}\ \min_{i\in\mathcal{L}}\ \mbox{\texttt{dist}}(\ell,i)$, and $R^{\max} = \max_{\ell\in\mathcal{I}}\ \max_{i\in\mathcal{L}}\ \mbox{\texttt{dist}}(\ell,i)$. Here, $\mbox{\texttt{dist}}(\ell,i)$ denotes the distance between locations $\ell$ and $i$ according to an underlying transportation network.\footnote{The lower bound for $r$ stems from the observation that when $r<R^{\min}$, even if charging stations are placed at all candidate sites, there would be at least one uncovered location of interest; so, no feasible solution exists for such $r$. The upper bound for $r$ follows from the fact that even if there is only a single selected site where a charging station is placed, and the solution is feasible, $R^{\max}$ is, by definition, the maximum distance to be travelled in order to reach that charging station.} Let $D^*(r)$ denote the maximum demand covered (obtained from a solution to the MPC problem (Figure \ref{fig:eqn-MPC}, perhaps by using Algorithm~\ref{alg:iter-pack-cover}) for a given reachability radius $r$. (For convenience, we define $D^*(r)=0$ for those $r$ for which MPC is infeasible.) If $\alpha\in[0,1]$ is a given trade-off parameter, then the objective can be defined as:
\begin{equation}\label{reachability}
\max_{r\in[R^{\min},R^{\max}]} \alpha\frac{D^*(r)}{\sum_{i\in \mathcal{L}} d_i} + (1-\alpha)\frac{R^{\max}-r}{R^{\max}-R^{\min}}.
\end{equation}
Let $\mathcal{R}=\{\mbox{\texttt{dist}}(\ell,i)|\ell\in\mathcal{I},i\in\mathcal{L},\mbox{\texttt{dist}}(\ell,i)\geq R^{\min}\}$ denote the set of all distances from a location of interest to a candidate site that is at least $R^{\min}$, and suppose we represent this set as $\mathcal{R}=\{r_1,r_2,\ldots,r_{|\mathcal{R}|}\}$, where $R^{\min}=r_1<r_2<\ldots<r_{|\mathcal{R}|}=R^{\max}$. The following lemma presents two observations concerning $D^*(r)$:
\begin{lemma}
$D^*(r)$ is nondecreasing,\footnote{This is because, for any two radii $r_1$, $r_2$ with $r_1\leq r_2$, the feasible set of MPC for $r_1$ is a subset of that for $r_2$.} and for all $1\leq m < |\mathcal{R}|$, for all $r_m\leq r < r_{m+1}$, $D^*(r)=D^*(r_m)$.
\end{lemma}
In other words, $D^*(r)$ is a nondecreasing step function and remains unchanged between values of $r$ that do not correspond to $\mbox{\texttt{dist}}(\ell,i)$ for some $(\ell,i)$. Thus, it can be seen that the value(s) of $r$ at which the objective function in \eqref{reachability} attains its maximum must be among the elements in $\mathcal{R}$. Therefore, solving \eqref{reachability} is equivalent to solving:
\begin{equation}\label{reachability-integral}
\max_{r\in\mathcal{R}} \alpha\frac{D^*(r)}{\sum_{i\in \mathcal{L}} d_i} + (1-\alpha)\frac{R^{\max}-r}{R^{\max}-R^{\min}}.
\end{equation}

\subsection{Experimental results.}\label{ssec:expt-optimal-placement}

We conducted two sets of experiments to evaluate the performance of IPAC using real world data -- one for North East England (presented in this section), and the other for the western region of East Anglia and southern region of East Midland (presented in Appendix~\ref{appendix-second-expt}).

\subsubsection{Setup.}
We use 1305 parking locations in North East England for both the candidate sites for placing charging stations, as well as the locations of interest that need to be covered. This is equivalent to a requirement that each candidate site must either have a charging station on-site, or one that is a short driving distance away. Accordingly, the cover set $S_i^r$ for each candidate site $i$ consists of the candidate sites that are within a distance of $r$ from $i$. The charging demands at candidate sites are predicted using the MDR model trained in Section \ref{ssec:expt-demand-prediction}. The costs are computed as $c_i = N_i(L_i+F_i)$; we estimate each of these components as follows:
\begin{enumerate}
\item \textit{$L_i$, the per-unit land cost}: Points of interest around candidate sites play vital roles in determining their land values. In addition, different types of facilities affect the land value differently. Thus, we assume $L_i = p + \sum_{j\in P_i^\delta}\frac{\mbox{\texttt{Score}}_{ij}}{\mbox{\texttt{dist}}(i,j)}$, where $p=\$4000$ is the minimum per-unit land cost across all the locations in North East England (computed using data obtained from~\cite{NEUK-Land-Value}), $P_i^\delta$ is the set of points of interest that are within a radius $\delta=1$~km from candidate site $i$, and $\mbox{\texttt{Score}}_{ij}$ is the score assigned to the point of interest $j$ according to its type, as follows. Airports and railway stations have the highest score of $800$, whereas schools, restaurants and hospitals have a lower score of $300$. $\mbox{\texttt{Score}}_{ij}$ is then normalized by $\mbox{\texttt{dist}}(i,j)$.
\item \textit{$F_i$, the per-unit infrastructure cost}: For Level 2 charging, we set $F_i=\$1852$ from Table~6 of~\cite{EVCS-Cost-Luskin}.
\item \textit{$N_i$, the number of charging spots}: We assumed a Level~2 charging rate of $6.4\mbox{kW}$ and set, for each candidate site, $N_i$ to be the minimum number of Level~2 charging spots necessary (using the queueing model described in Section~\ref{appendix-cost-queueing}) to ensure that the average ``peak-demand'' waiting time (taken as the estimated maximum hourly demand at the candidate site over two years) is less than $5$ minutes.
\end{enumerate}
The predicted demands and estimated costs of the candidate sites are then used to find the optimal locations using Algorithm~\ref{alg:iter-pack-cover} (IPAC), where, for both \texttt{KP()} and \texttt{SC()}, we choose the well-known greedy approximation algorithms introduced in~\cite{Vazirani01}, and for \texttt{RANK()}, we use the function proposed in Section~\ref{iter-pack-cover}.

\begin{figure*}[!ht]
    \centering
    \subfigure[\scriptsize{Demand satisfied by LP-relaxation, Optimal ILP-solution, IPAC-heuristic, and Naive-heuristic.}]{\label{fig:expt1a}\includegraphics[height=1.5in, width=0.32\textwidth]{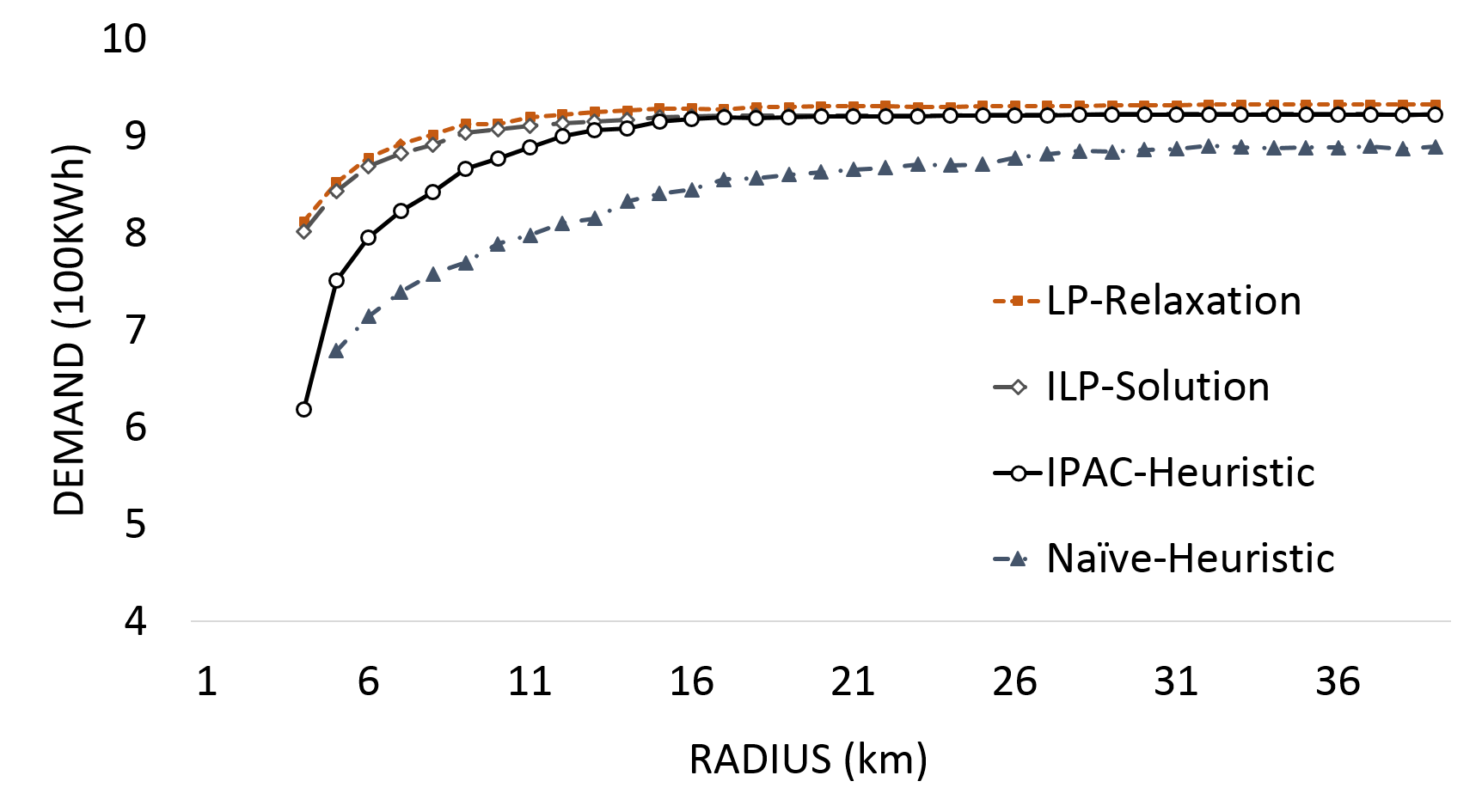}}
    \hfill
    \subfigure[\scriptsize{Fraction of ILP-solution demand satisfied by IPAC-heuristic and Naive-heuristic.}]{\label{fig:expt1b}\includegraphics[height=1.5in, width=0.32\textwidth]{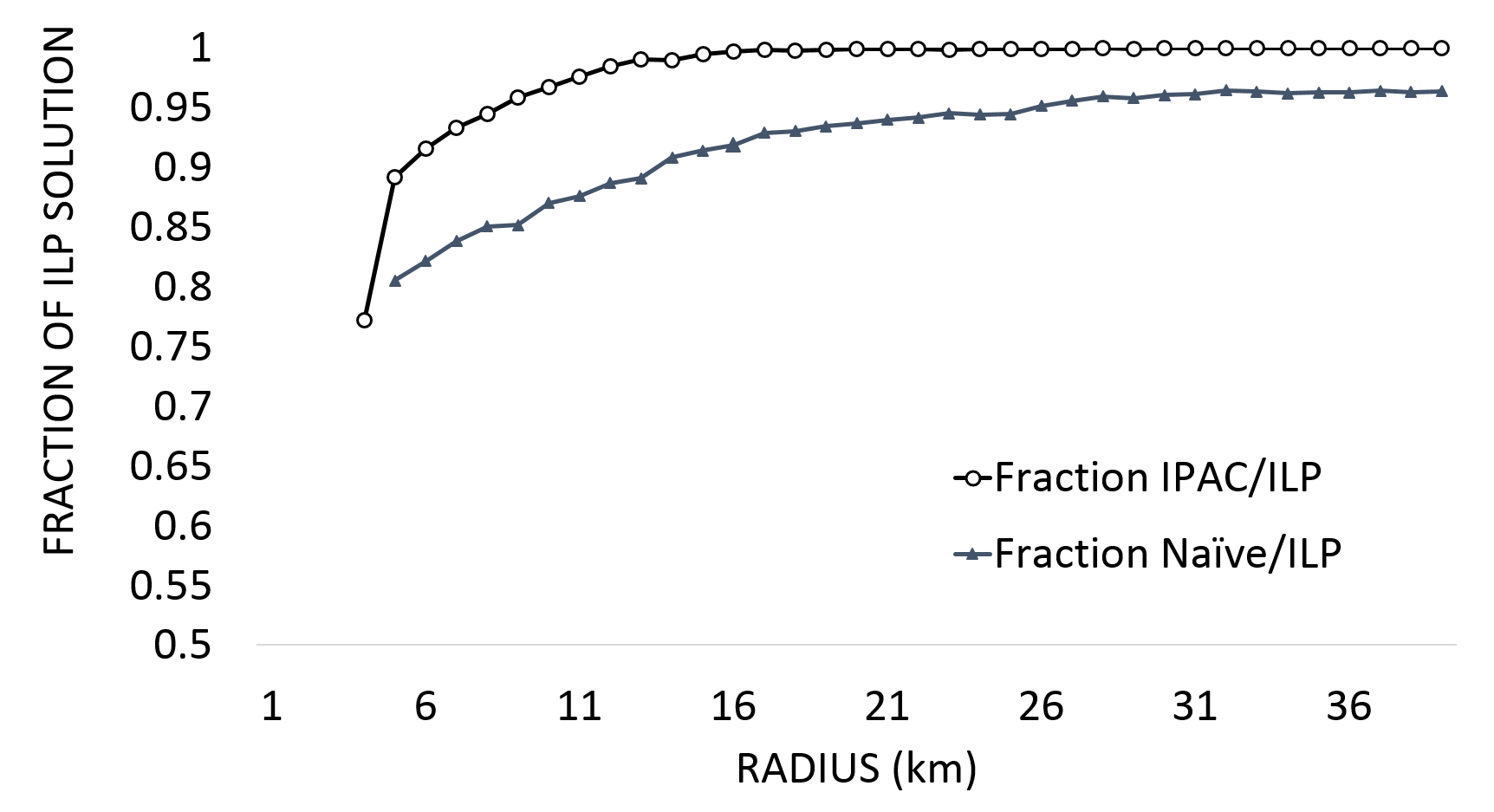}}
    \hfill
    \subfigure[\scriptsize{Minimum required budget for feasibility by IPAC-heuristic and LP-relaxation.}]{\label{fig:expt2a}\includegraphics[height=1.5in, width=0.32\textwidth]{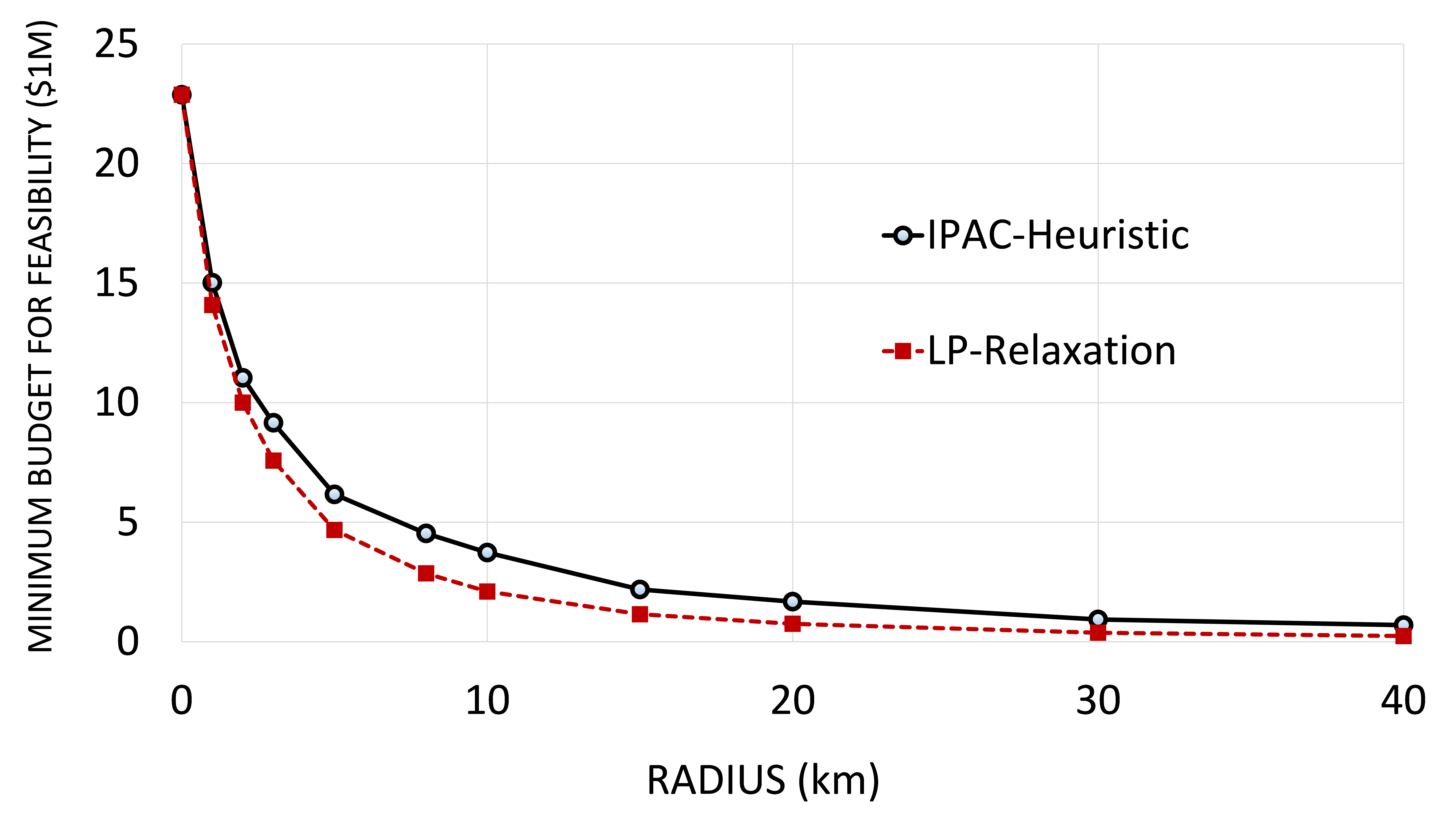}}
    \\
    \subfigure[\scriptsize{Increase in demand satisfied by IPAC-heuristic over Naive-heuristic.}]{\label{fig:expt1e}\includegraphics[height=1.5in, width=0.49\textwidth]{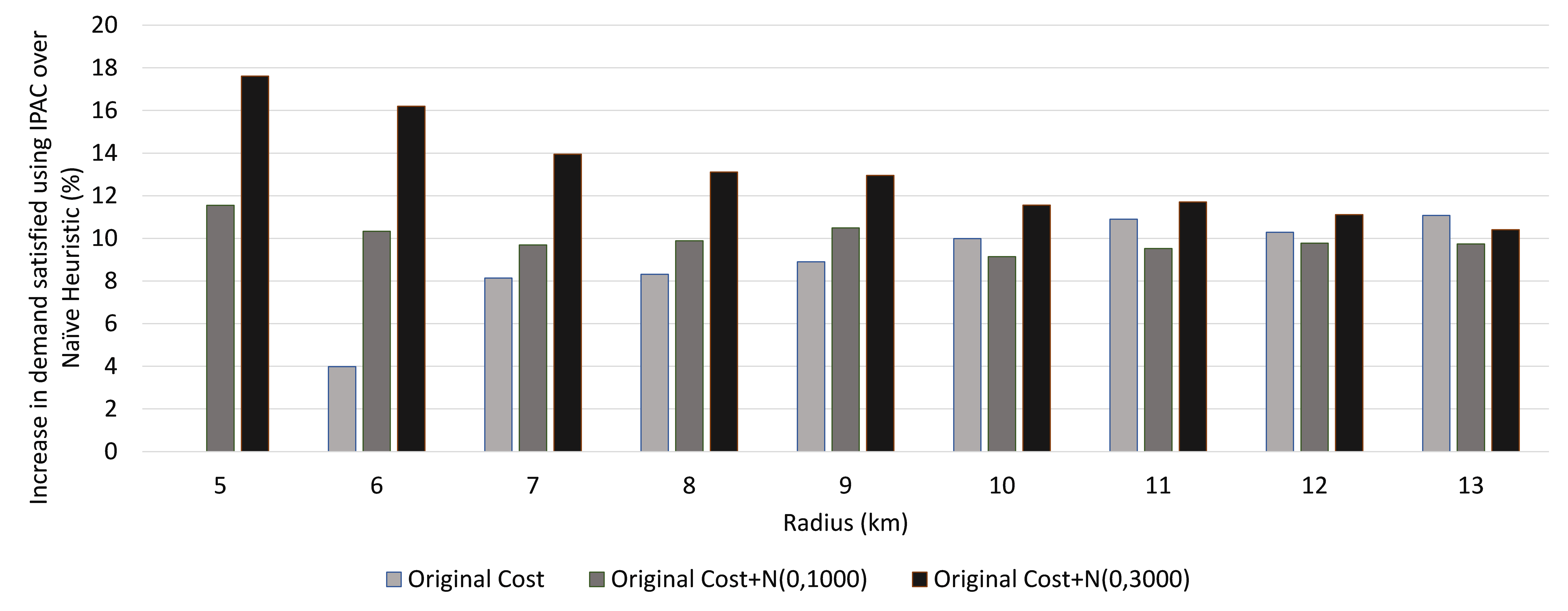}}
    \hfill
    \subfigure[\scriptsize{Feasibility gap between IPAC-heuristic and Naive-heuristic.}]{\label{fig:expt1d}\includegraphics[height=1.5in, width=0.49\textwidth]{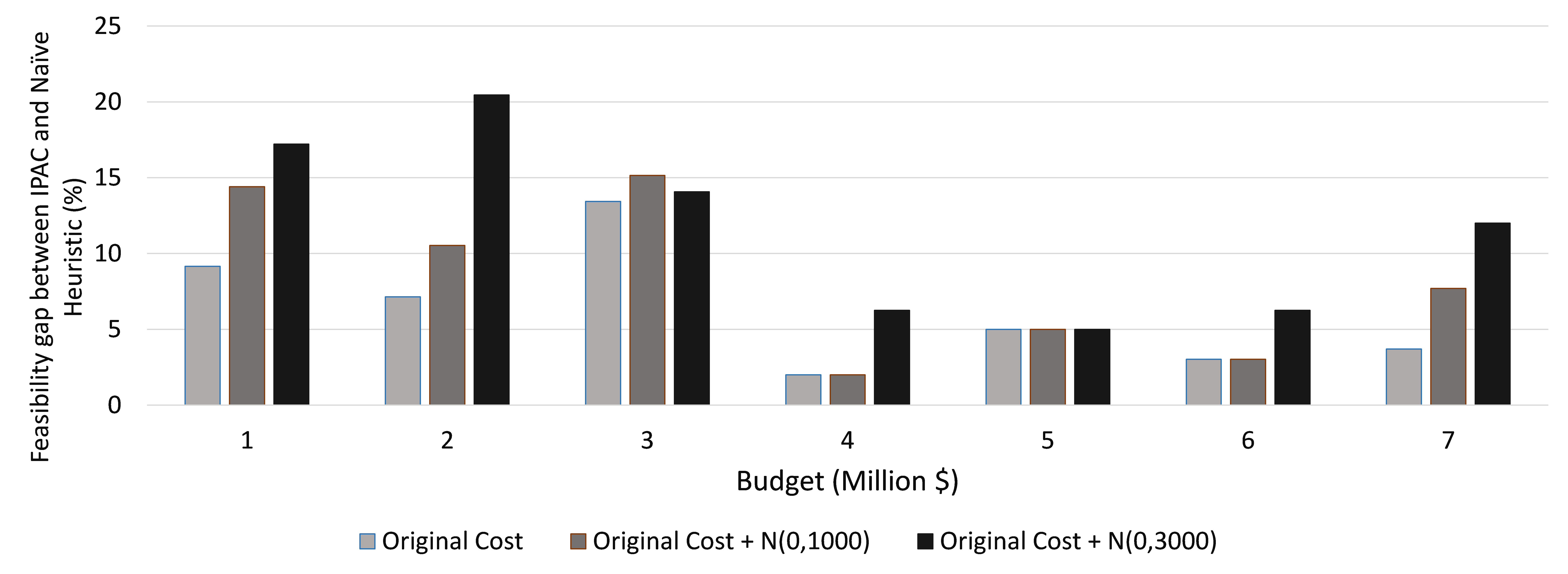}}
    \\
    \subfigure[\scriptsize{Running time of IPAC as compared to directly solving the ILP using CPLEX.}]{\label{fig:expt-runningtime}\includegraphics[width=0.45\textwidth]{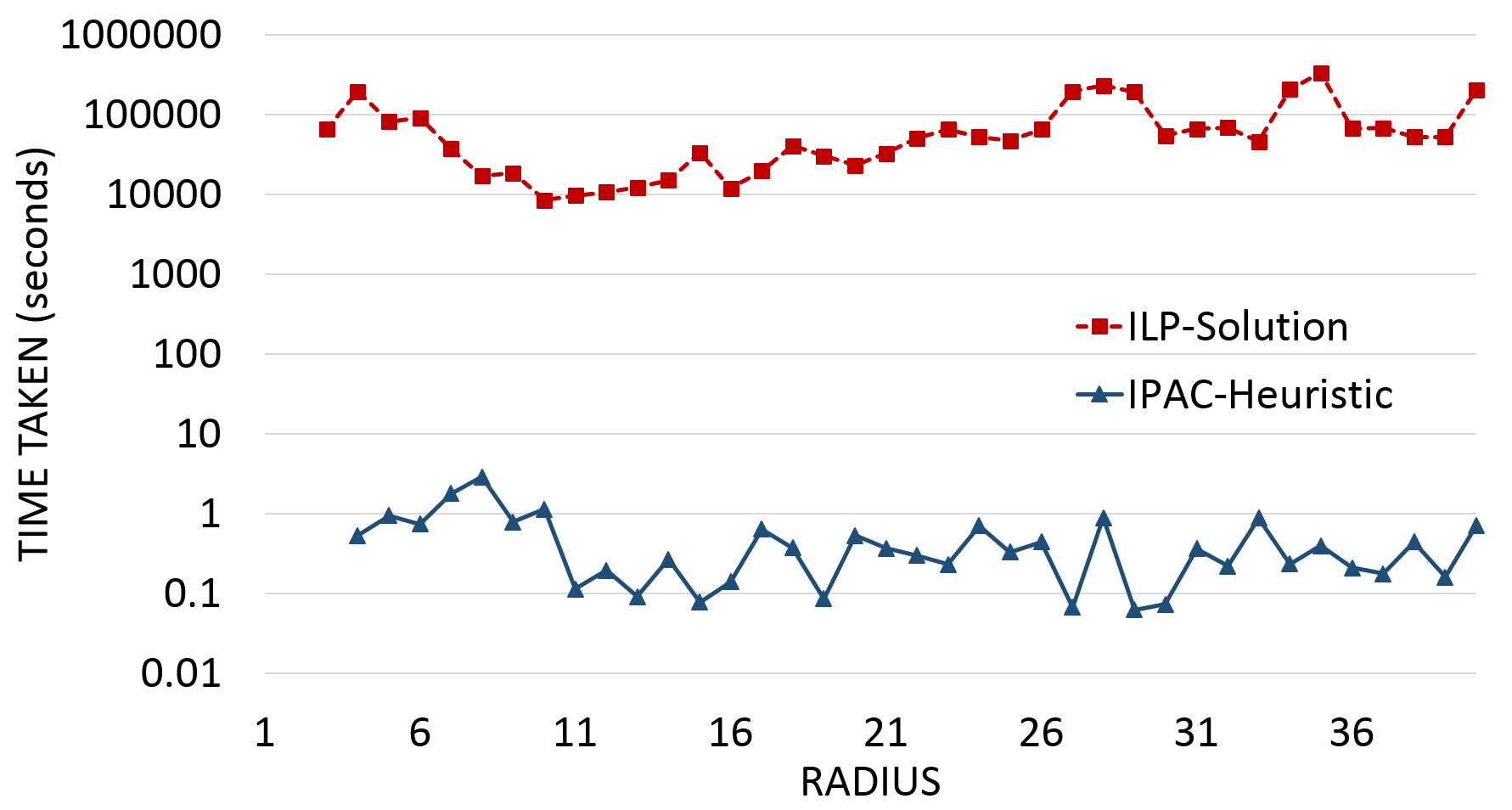}}
    \caption{Experimental results for North East England.\label{fig:expt1}}
\end{figure*}

\subsubsection{Performance.}
To evaluate the performance of IPAC when it can locate feasible solutions, we compare it with (i)~a naive heuristic that first solves the covering problem using \texttt{SC()}, and then invokes \texttt{KP()} on the remaining budget to add unselected candidate sites, (ii)~the optimal ILP solution of the MPC problem  (Figure~\ref{fig:eqn-MPC}), and (iii)~a solution to the LP relaxation of MPC problem (Figure~\ref{fig:eqn-MPC}), which gives an upper bound on the actual (integer LP) optimization problem. We use a budget of $B=\$6\mbox{M}$. The number of pairwise distances between candidate sites is prohibitively large for our experiment; so, we used values of $r$ (in units of km), from the set $\{5.1,6,7,8,9,10,12,15,20,25,30,40\}$. The results are plotted in Figures~\ref{fig:expt1a}-\ref{fig:expt1b}.

As we increase the reachability radius $r$, the set of allocations that satisfy the covering constraints steadily gets larger, until, for a large enough radius, any allocation would satisfy the covering constraints, reducing the optimization to a pure packing problem. Since the feasibility set only gets larger with $r$, the demand covered should also increase accordingly. The graphs validate this expected behavior. In addition, we observe that the demand captured by IPAC is far closer to that of the LP relaxation than that captured by the naive heuristic. In particular, from Figure~\ref{fig:expt1b}, we can see that when $r=9\mbox{km}$, the IPAC heuristic already captures almost $90\%$ of the LP relaxation demand and its performance is always better than the naive heuristic. Also, the time taken to obtain the ILP solution for a given radius using CPLEX~\cite{cplex2009},\footnote{\scriptsize{Intel(R) Xeon (R) CPU @ 2.2 GHz (16 cores), 32GB RAM, 64-bit Windows.}} ranges from $23-30$ hours for most instances, whereas IPAC finished in less than 0.5 seconds; see Figure~\ref{fig:expt-runningtime}.

\subsubsection{Feasibility.}
Since IPAC is a heuristic, it may falsely deem an instance of the MPC problem (Figure~\ref{fig:eqn-MPC}) infeasible, when in fact, feasible solutions exist. Thus, it is important to first analyze the extent of this limitation. It is easy to observe that an instance of MPC is feasible if and only if the available budget $B$ is greater than or equal to the minimum required budget determined by solving the corresponding instance of the set-cover subproblem (Figure~\ref{fig:eqn-SC}). In the case of IPAC (Algorithm~\ref{iter-pack-cover}), it reduces to whether the available budget $B$ is greater than or equal to the minimum required budget as determined by \texttt{SC()}.

Therefore, given an instance of MPC, we look at the corresponding set-cover subproblem and compare its solutions (the minimum required budgets) as obtained by (i)~using the greedy approximation algorithm in~\cite{Vazirani01}, and (ii)~directly solving its relaxed LP.\footnote{\scriptsize{Computing ILP solutions for feasibility analysis exactly is prohibitively time-consuming, since it must be computed for various $r$ and $B$ values.}} We use different values of $r$ (in units of km) to generate instances. The results are plotted in Figure~\ref{fig:expt2a}. (Since the LP relaxation allows for fractional allocations, the corresponding solution is only a lower bound on the minimum budget required for feasibility.)

\subsubsection{Accounting for noisy costs.}
Since our model for estimating the costs may not be accurate, we perform experiments for two additional scenarios, where we add zero-mean gaussian noise to the costs --- one with a standard deviation of \$1000 and another with a standard deviation of \$3000. In each case, we compare IPAC and the naive heuristic:
\begin{itemize}
\item \textit{Performance:} Using a budget of \$6M, we calculate the difference in the demands satisfied by the solution obtained using IPAC and the naive heuristic. Figure~\ref{fig:expt1e} shows this difference as a percentage, for several values of the reachability radius. It can be seen, once again, that IPAC's advantage is significant (8-18\% in most cases).
\item \textit{Feasibility:} We calculate the minimum reachability radius for which IPAC and the naive heuristic are able to find a feasible solution to MPC, and find that IPAC is always able to find feasible solutions for smaller radii. Figure~\ref{fig:expt1d} shows this difference as a percentage, for several values of the available budget. It can be seen that IPAC's advantage over the naive heuristic is particularly remarkable (10-20\% in most cases) when the budget is scarce.
\end{itemize}

Thus, experimental results show that IPAC's feasibility gap as compared to the LP-Relaxation is not significant, and IPAC's performance quickly approaches that of the LP-Relaxation demand. Further, in terms of both feasibility and performance, it can be seen that the advantage of IPAC over the naive heuristic is significant (10-20\% in most cases) and this advantage is fairly robust to noise in the estimated costs.

\section{Extensions and future work.}

In this section, we explain how the optimization framework we introduced in Section \ref{sec:placement-optimization} can be extended to a situation where the deployment of charging stations is incremental, and conclude with some open questions.\footnote{Even though the work in this paper is motivated by EV charging stations, the problem formulation, modeling and solution frameworks presented here can be extended to other facility location problems which have demand, budget and coverage requirements, such as placement of bus-stop shelters, parking lots, and healthcare kiosks.} To highlight another important alternate scenario, consider a set of private charging station providers that would like to set up charging stations, but would need government subsidies to incentivize them to do so, especially in regions of low demand. The government agency would then run a grant program~\cite{Grants-CA,Grants-UK}, where the providers specify the subsidies they need to set up charging stations at candidate sites they are interested in. The government has a budget from which to allocate grants; moreover, the providers come with their own budget constraints. In Appendix \ref{ssec:government-grant}, we expand the placement optimization problem (MPC) to accommodate these additional packing constraints and explain how to extend the IPAC framework accordingly by solving a multi-dimensional knapsack subproblem.

\subsection{Incremental (multi-period) placement.}\label{ssec:incremental}

In practice, the budget for the deployment of charging stations may be released over time, and the demands at candidate sites and charging station installation costs may change over time. We now extend the problem formulation and the heuristic framework presented in Section \ref{sec:placement-optimization} for this case. Let there be $T$ time periods, $t = 1, \ldots, T$. We first extend earlier notation to include time periods, by superscripting them with $t$. Then, the single-period formulation in Figure~\ref{fig:eqn-MPC} and Equation~\eqref{reachability-integral} can be combined and generalized for the multi-period placement optimization problem as follows:
\begin{equation}\label{mpct}
\begin{split}
\max_{\substack{r^t\in[R^{\min},R^{\max}]\\ x^{t}_{i}\in\{0,1\}}} &\sum_{t=1}^{T} \left( \alpha\frac{\sum_{i\in\mathcal{L}^t}d^{t}_{i} x^{t}_{i}}{\sum_{i\in \mathcal{L}^t} d^{t}_{i}} + (1-\alpha)\frac{R^{\max}-r^t}{R^{\max}-R^{\min}} \right) \\
\mbox{subject to } &x^{t}_{i} \geq x^{t-1}_{i} \qquad \forall\ i \in\mathcal{L}^{t-1}; 1 \leq t \leq T \\
&\sum_{t=1}^{\tau} \sum_{i\in\mathcal{L}^t}c^{t}_{i} (x^{t}_{i} - x^{t-1}_{i}) \leq \sum_{t=1}^{\tau} B^t \quad \forall\ 1 \leq \tau \leq T\\
&\sum_{i\in\mathcal{L}:\ell\in S_{i}^{r^{t}}}x^{t}_{i} \geq 1 \qquad \forall\ \ell\in\mathcal{I}^t; 1 \leq t \leq T\\
\end{split}
\end{equation}
In the above formulation, initially, $x^{0}_{i}=0$ at all candidate sites $i\in\mathcal{L}^0$. If a charging station is placed at location $i$ at time $t>0$, then $x^{t}_{i}$ is set to $1$, and the first constraint ensures that the charging station remains installed for all subsequent periods. The second constraint ensures that the cost of all (new) installations done at time $\tau$ is within the budget released at $\tau$ plus any leftover budget from previous time periods, and the third set of constraints is the covering constraints for each time period $t$, as a function of the reachability radius $r^t$. The objective is to maximize the fraction of demand satisfied while minimizing the reachability radius, summed over all time periods. A straightforward ``greedy'' heuristic for this problem is to predict demands and estimate costs at the remaining candidate sites at the beginning of each time period $t$, and then using Algorithm \ref{alg:iter-pack-cover} (IPAC) to locate additional charging station installations.

A common and important scenario that any solution for incremental placement must handle is future increase in demand at previously deployed charging stations. The greedy heuristic above only indirectly addresses this concern by ensuring that the increased demand at such charging stations would be taken into account by the demand prediction algorithm in subsequent time periods. Still, a simple modification can allow for the option of expanding existing charging stations by adding more charging slots in response to increasing demand---let the set of candidate sites at time period $t+1$ be a superset of the set of candidate sites at time period $t$, i.e., $\mathcal{L}^{t+1}\supseteq\mathcal{L}^t$, but the costs $c^{t+1}_i$ at previously selected candidate sites $i$ would now be the cost of adding additional charging slots to maintain the waiting time constraints. This would let IPAC decide whether the right thing to do in response to increasing demand is to expand an existing charging station or set up a new charging station nearby. While our greedy heuristic is a starting point, more effective algorithms likely exist, especially when future demands and installation costs can be predicted with reasonable accuracy. We leave such directions to future work.

\subsection{Extending the IPAC framework.}

Further analysis of the class of algorithms defined by the IPAC framework would be of interest. In particular, it is worth exploring provable performance guarantees of an IPAC algorithm, e.g., Algorithm \ref{alg:iter-pack-cover}, in terms of those of its constituent knapsack and set-cover algorithms. Another interesting direction would be to investigate if the demand prediction model can be integrated into the optimization framework for the placement of charging stations, since the placement of a charging station at one candidate site would likely affect the demand at nearby candidate sites.



\bibliographystyle{ormsv080}
\bibliography{submission-placement}

\newpage

\begin{APPENDICES}

\section{Approximation algorithm for a transformed problem.}\label{appendix-approx-algo}
This section is an expanded version of the discussion towards the end of Section \ref{ssec:prelims-placement-optimization} which includes a proof of Lemma~\ref{lemma:appendix-approx-algo}. We present an approximation algorithm for a pure covering problem that is closely related to the Mixed Pack \& Cover (MPC) problem (Figure \ref{fig:eqn-MPC}), which serves as a basis for designing a heuristic framework for solving MPC.

Consider the following Demand and Set-Cover (DSC) problem, which does not contain any packing constraint, but is closely related to the MPC problem:

\begin{small}
\begin{equation}\label{dsc}
\begin{split}
\min_{\left\{x_i:i\in\mathcal{L}\right\}} &\sum_{i\in\mathcal{L}}c_i x_i\\
\mbox{subject to } &\sum_{i\in\mathcal{L}:\ell\in S_i^r}x_i \geq 1\qquad\forall\ \ell\in\mathcal{I}\\
&  \sum_{i\in\mathcal{L}}d_i x_i \geq D
\end{split}
\end{equation}
\end{small}

Intuitively, an algorithm for DSC can be instantiated with different values of total demand $D$, to search for a solution for MPC, e.g., with increasing values of total demand $D$ (between $\min_{d_i}$ and $\sum d_i$) until the total cost exceeds $B$. The IPAC framework for MPC presented in Section \ref{iter-pack-cover} is inspired by this observation. Now, to develop an approximation algorithm for DSC, consider its two subproblems:

\begin{small}
\begin{equation}\label{dsc-sc}
\begin{split}
\min_{\left\{x_i:i\in\mathcal{L}\right\}} &\sum_{i\in\mathcal{L}}c_i x_i\\
\mbox{subject to } &\sum_{i\in\mathcal{L}:\ell\in S_i^r}x_i \geq 1\qquad\forall\ \ell\in\mathcal{I}
\end{split}
\end{equation}
\end{small}

\begin{small}
\begin{equation}\label{dsc-minKP}
\begin{split}
\min_{\left\{x_i:i\in\mathcal{L}\right\}} &\sum_{i\in\mathcal{L}}c_i x_i\\
\mbox{subject to } &\sum_{i\in\mathcal{L}}d_i x_i \geq D
\end{split}
\end{equation}
\end{small}

Problems~\eqref{dsc-sc} and~\eqref{dsc-minKP} are the well-known NP-Hard problems Set-Cover (SC)~\cite{Vazirani01} and Minimization Knapsack (MinKP)~\cite{Csirik91}, respectively. Let, $OPT_{dsc}$, $OPT_{sc}$ and $OPT_{MinKP}$ be the costs of an optimal solution of problems~\eqref{dsc},~\eqref{dsc-sc}, and~\eqref{dsc-minKP}, respectively. Since every feasible solution for problem~\eqref{dsc} is a feasible solution for problem~\eqref{dsc-sc} as well as problem~\eqref{dsc-minKP}, $OPT_{sc} \leq OPT_{dsc}$ and $OPT_{MinKP} \leq OPT_{dsc}$.

Consider an $f_{sc}$ factor approximation algorithm $A_{sc}$ for set-cover and $f_{kp}$ factor approximation algorithm $A_{kp}$ for minimization knapsack. Let, $\mathcal{L}_{sc}$ and $\mathcal{L}_{kp}$ (subsets of $\mathcal{L}$) be the solutions returned by $A_{sc}$ and $A_{kp}$, respectively. Consider algorithm $A_{dsc}$ for DSC that uses $A_{sc}$ and $A_{kp}$ as subroutines, and returns the union of the solutions returned by the two subroutines, i.e., $\mathcal{L}_{sc} \cup \mathcal{L}_{kp}$, for problem~\eqref{dsc}. Clearly, $\mathcal{L}_{sc} \cup \mathcal{L}_{kp}$ is a feasible solution for DSC since locations in $\mathcal{L}_{sc}$ help satisfy the set-cover constraints, and locations in $\mathcal{L}_{kp}$ help satisfy the demand-cover constraint. Moreover, $c(\mathcal{L}_{sc} \cup \mathcal{L}_{kp}) \leq c(\mathcal{L}_{sc}) + c(\mathcal{L}_{kp}) \leq  f_{sc}.OPT_{sc} + f_{kp}.OPT_{MinKP} \leq (f_{sc}+f_{kp}).OPT_{dsc}$. Here $c(\cdot)$ denotes the cost of a solution, that is, the sum of the costs of all locations that appear in a solution.

Thus, $A_{dsc}$ is an ($f_{sc}+f_{kp}$) factor approximation algorithms for DSC. Substituting $A_{sc}$ and $A_{kp}$ with  known set cover and minimization knapsack algorithms from literature gives different approximation algorithms for DSC. For example, using the $f$-factor algorithm for set-cover from~\cite{Vazirani01} (where $f$ is the frequency of the most frequent element, i.e., maximum number of reachability sets in which a candidate site appears), and the $2$~factor algorithm for minimization knapsack from~\cite{Csirik91}, gives an $(f+2)$ factor algorithm for DSC.

\section{Alternate scenario: Granting subsidies to charging station providers.}\label{ssec:government-grant}

The problem modeled in Section \ref{sec:placement-optimization} and its extension in Section \ref{ssec:incremental} focused on a situation in which a central planner such as a government agency would use funds to construct charging stations in a geographical region. Hence, $c_i$ referred to the cost to the government agency of setting up a charging station at $i$. An alternate scenario could be one where there is a set $\mathcal{P}=\{1,2,\ldots,|\mathcal{P}|\}$ of private charging station providers that would like to set up charging stations, but would need government subsidies to incentivize them to do so, especially in regions of low demand. In this case, the government agency runs a grant program, where the providers specify the subsidies they need to set up charging stations at candidate sites they are interested in. Instead of $c_i$, we now have $c_{ij}$ denoting the subsidy to be paid to provider $j\in\mathcal{P}$ at candidate site $i\in\mathcal{L}$, if selected.

We solicit the following additional inputs, from the private providers:\footnote{The procedure usually involves a screening stage where the proposals submitted by the providers are evaluated for the accuracy/legitimacy of their cost estimation and project feasibility by a grant committee in order to determine their eligibility for further consideration.} (a)~preferences regarding which candidate sites the providers are interested in setting up and operating charging stations, (b)~estimates of costs and budgets, and (c)~bids on desired subsidies. We then determine the subset of candidate sites selected for deployment, the ``winning'' providers for each of the selected sites, and the subsidy amounts allocated to each of them. In addition to demand maximization, coverage and government agency budget constraints, the additional constraints that should be taken into consideration are the site preferences and budget constraints of the providers.

\subsection{Optimal placement with subsidy allocation.}

A couple of concerns arise in allowing multiple private providers to compete for setting up and operating charging stations:
\begin{itemize}
\item Providers can express interest in multiple sites, even though they may not have the means to build charging stations at all those sites. Each provider $j\in\mathcal{P}$ would therefore also specify their budget $B_j$, and their estimated costs $p_{ij}$ (after taking into account the subsidy $c_{ij}$) of building a charging station at candidate site $i\in\mathcal{L}$. It is then the government agency's responsibility to ensure that the outcome does not violate any provider's budget constraint. This adds additional \textit{packing} constraints to the problem. Note that we allow for providers to specify \textit{site-specific} subsidies, because the amount of money they are willing to spend setting up a charging station at a given site is likely to depend on that site's demand.
\item There could be candidate sites (e.g., with relatively low demand) that no provider is interested in, or, even if there is interest, providers might ask for very high subsidies, so that even the least of these would cost the government agency more than what it would cost to build a charging station by itself. Furthermore, it is likely that some such sites with low demand are the only ones that cover a remote area, one of which must therefore be selected in any feasible allocation. To handle such scenarios, we model the government agency to be a ``provider'' as well, who, for every site, asks for a subsidy of $c_i$, which is the cost to the government agency of constructing its own charging station at that site.\footnote{This is equivalent to setting a \textit{reserve} subsidy at each site.} Therefore, we add the government agency to the set of providers $\mathcal{P}$ and denote it with the index $0$, to obtain the set of participants, $\mathcal{B}=\{0,1,2,\ldots,|\mathcal{P}|\}$.
\end{itemize}
To recap our notation for this scenario, $\mathcal{L}=\{1,2,\ldots,|\mathcal{L}|\}$ denotes the set of all candidate sites for charging stations, $r\in\mathbb{R}_+$ denotes the reachability radius, $\mathcal{I}$ denotes the set of all locations of interest to be covered, for each candidate site $i\in\mathcal{L}$, $d_i\in\mathbb{R}_+$ is the demand for charging at $i$, and $S_i^r\subseteq\mathcal{I}$ is the set of locations of interest that would be covered if a charging station is placed at $i$.

Let $\mathcal{P}=\{1,2,\ldots,|\mathcal{P}|\}$ denote the set of charging station providers. For each provider $j\in\mathcal{P}$,
\begin{enumerate}
\item $B_j\in\mathbb{R}_+$ is the total budget available,
\item $c_{ij}\in\mathbb{R}_+$ is the subsidy that the provider specifies they need at site $i\in\mathcal{L}$, and
\item $p_{ij}\in\mathbb{R}_+$ is the cost to the provider of building a charging station at site $i\in\mathcal{L}$, after taking into account the subsidy $c_{ij}$.
\end{enumerate}
$\mathcal{B}=\mathcal{P}\cup\{0\}$ denotes the set of participants after adding the government agency to the set of providers. $B\in\mathbb{R}_+$ denotes the total budget available to the government agency, and $c_{i0}=c_i$ is the government agency's subsidy specification for site $i\in\mathcal{L}$. For each site $i\in\mathcal{L}$, each participant $j\in\mathcal{B}$, the outcome $x_{ij}\in\{0,1\}$ denotes whether participant $j$ ``won'' site $i$.

Given a set of prices, budgets and subsidy specifications from the participants, the outcome is determined by the following optimization problem:

\vspace{-0.1in}
\begin{small}
\begin{equation}\label{mmpc}
\begin{split}
\max_{\left\{x_{ij}:i\in\mathcal{L},j\in\mathcal{B}\right\}} &\sum_{i\in\mathcal{L}}d_i\sum_{j\in\mathcal{B}}x_{ij}\\
\mbox{subject to } &\sum_{i\in\mathcal{L}}\sum_{j\in\mathcal{B}}c_{ij} x_{ij} \leq B\\
&\sum_{i\in\mathcal{L}}p_{ij} x_{ij} \leq B_j\qquad\forall\ j\in\mathcal{P}\\
&\sum_{j\in\mathcal{B}}x_{ij} \leq 1\qquad\quad\;\;\;\forall\ i\in\mathcal{L}\\
&\sum_{k:\ell\in S_k^r}\sum_{j\in\mathcal{B}}x_{kj} \geq 1\;\;\;\ \forall\ \ell\in\mathcal{I}
\end{split}
\end{equation}
\end{small}

\vspace{-0.1in}
The IPAC framework (and its instantiation in Algorithm \ref{alg:iter-pack-cover}) presented in Section \ref{iter-pack-cover} can be extended by (a)~replacing calls to \texttt{KP()} with calls to \texttt{MULTIKP()}, a method that solves the multi-dimensional knapsack problem, and (b)~incorporating the multiple knapsack costs of an item into the value function used in \texttt{RANK()}, and (c)~within each iteration, reducing multiple knapsack budgets and making sure that items are removed until \textit{all} reduced budget constraints are satisfied.

\section{Second experiment for optimal placement: Western part of East Anglia and southern part of East Midland.}\label{appendix-second-expt}

We also used IPAC to find the locations for optimal placement of charging stations across the western region of East Anglia and southern region of East Midland, where there are very few existing charging stations. We use parking locations in these areas as candidate sites for placing charging stations. We predict the demand and estimate the cost for the candidate sites as we did for North East England and used a budget of $B=\$800,000$. To evaluate the performance IPAC with the naive heuristic and the LP relaxation, we used values of $r$ (in units of km) from the set $\{4,5,6,8,10,12,15,20,30\}$. The corresponding results are shown in Figure \ref{fig:expt4}. It can be seen that qualitatively, the results are very similar to those obtained for North East England (Section \ref{ssec:expt-optimal-placement}).

\begin{figure*}[!hb]
    \vspace{-0.1in}
    \centering
    \subfigure[\scriptsize{Demand satisfied by LP-relaxation, IPAC-heuristic, and Naive-heuristic.}]{\label{fig:expt4a}\includegraphics[height=1.5in, width=0.32\textwidth]{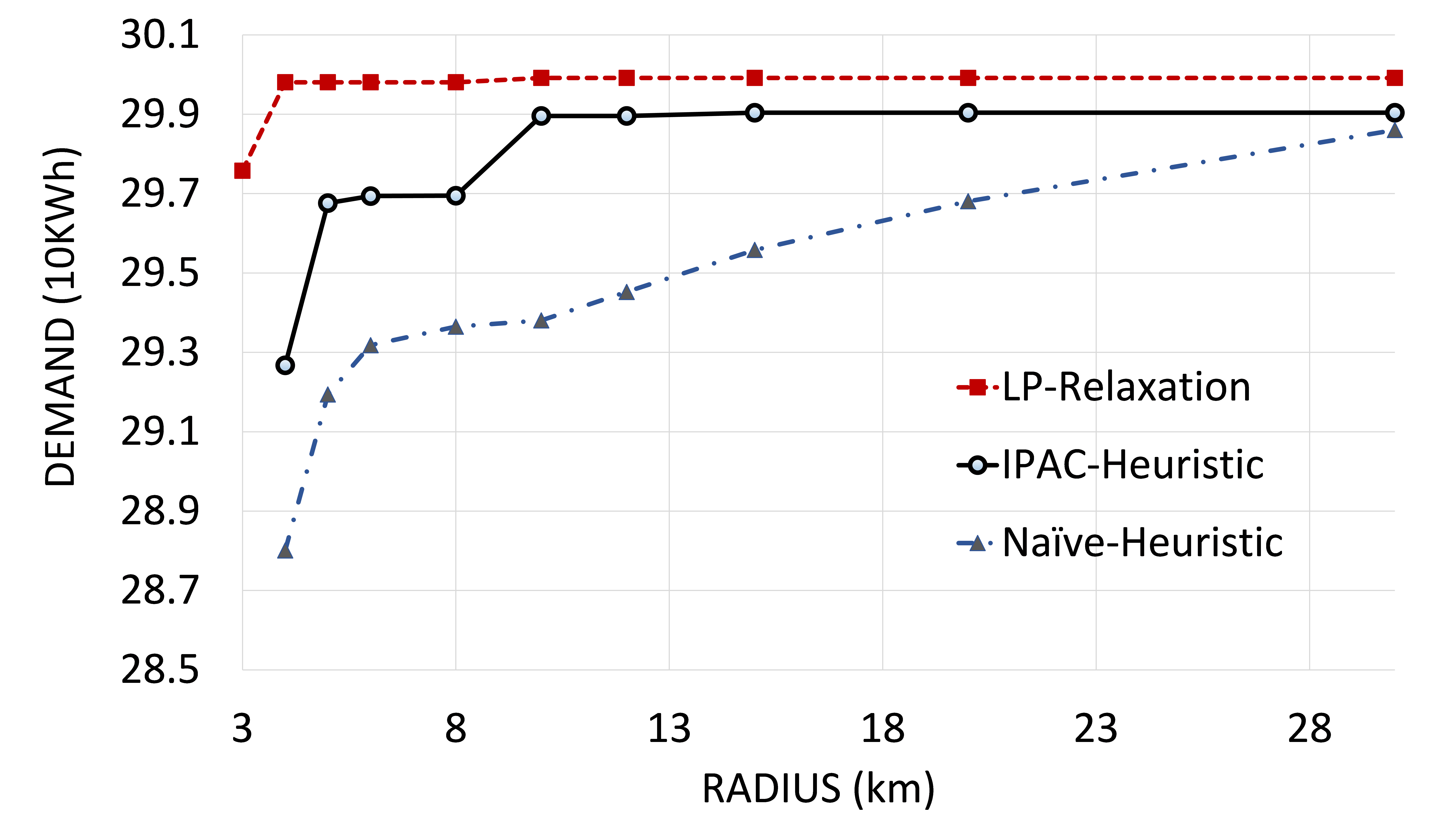}}
    \hfill
    \subfigure[\scriptsize{Fraction of ILP-solution demand satisfied by IPAC-heuristic and Naive-heuristic.}]{\label{fig:expt4b}\includegraphics[height=1.5in, width=0.32\textwidth]{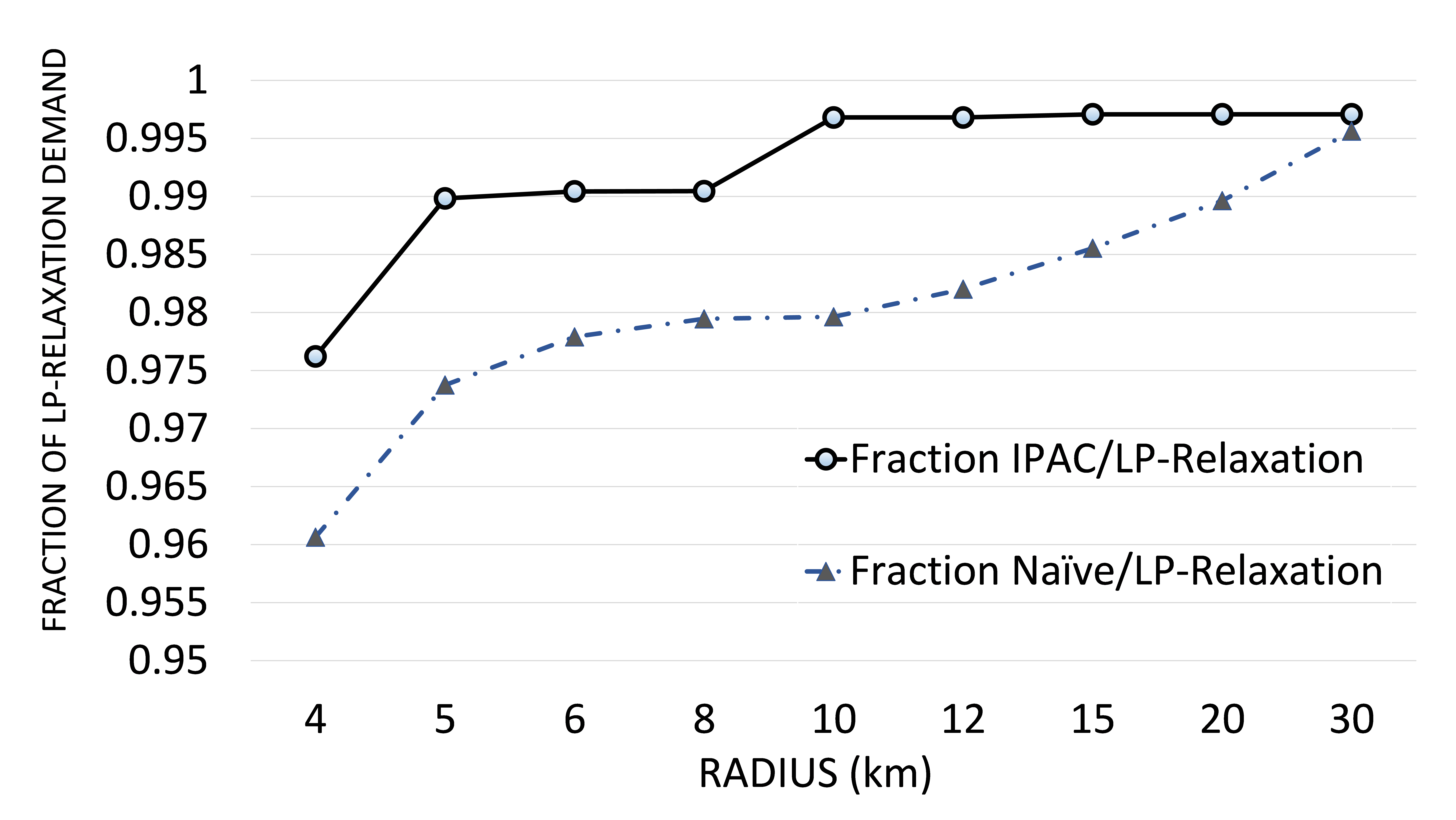}}
	\hfill
    \subfigure[\scriptsize{Minimum required budget for feasibility by IPAC-heuristic and LP-relaxation.}]{\label{fig:expt4c}\includegraphics[height=1.5in, width=0.32\textwidth]{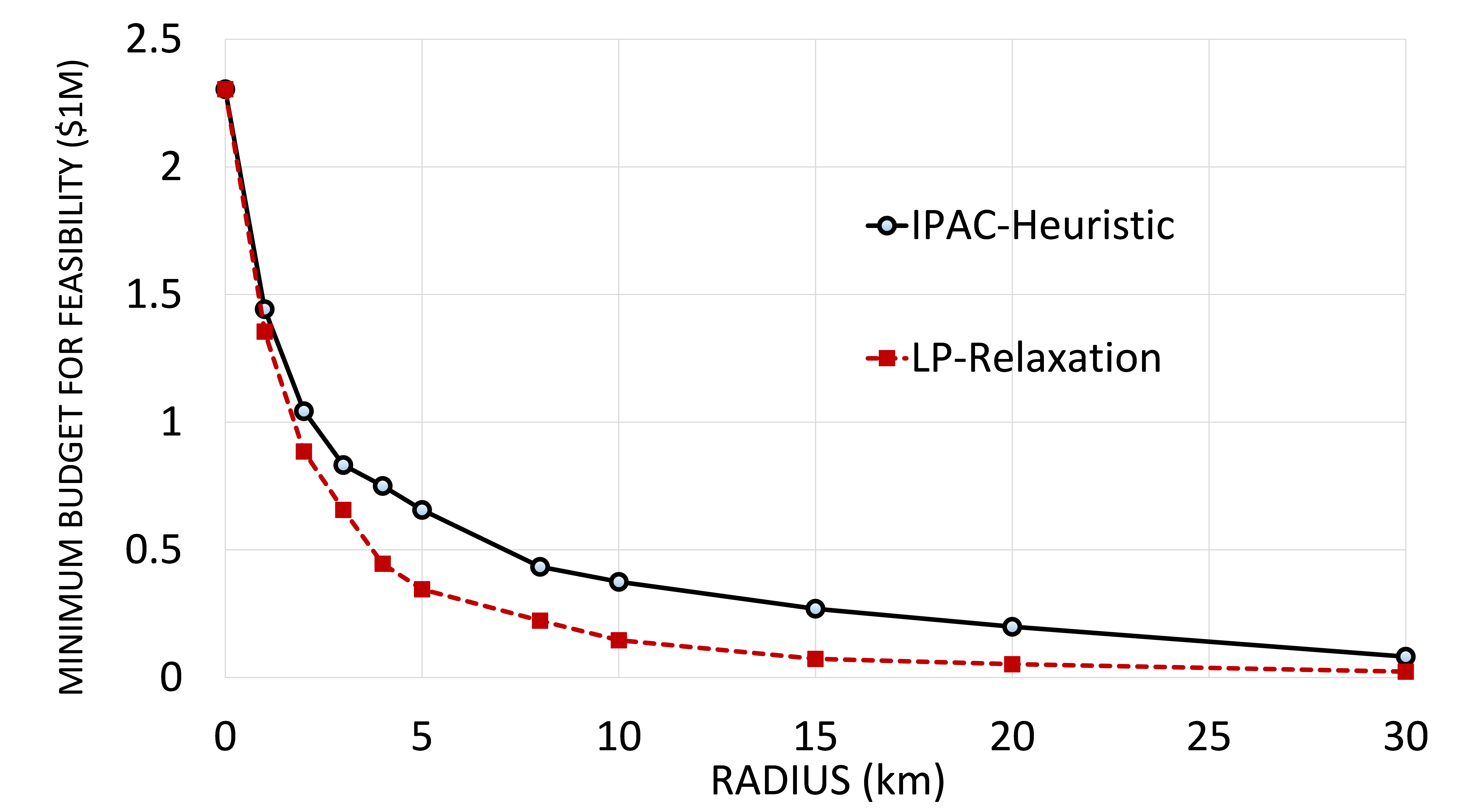}}
    \caption{Experimental results for western part of East Anglia and southern part of East Midland.\label{fig:expt4}}
\end{figure*}

\end{APPENDICES}

\end{document}